\documentclass[letterpaper]{article} 
\usepackage{aaai2027}  
\nocopyright
\usepackage[hyphens]{url}  
\usepackage{graphicx} 
\usepackage{natbib}  
\usepackage{caption} 
\usepackage{algorithm}

\usepackage{amsfonts} 
\usepackage{amsmath}
\usepackage{algpseudocode}
\usepackage{subcaption}
\usepackage{booktabs}

\usepackage{tabularx}
\usepackage{array}

\newcolumntype{Y}{>{\centering\arraybackslash}X}
\newcolumntype{C}[1]{>{\centering\arraybackslash}p{#1}}
\newtheorem{Theorem}{Theorem}
\usepackage{newfloat}
\usepackage{listings}
\DeclareCaptionStyle{ruled}{labelfont=normalfont,labelsep=colon,strut=off} 
\floatstyle{ruled}
\newfloat{listing}{tb}{lst}{}
\floatname{listing}{Listing}

\usepackage{booktabs}

\title{From Points to Edges: Edge-Conditioned Spectral Operators for Physics-Sensitive PDE Learning}

\author {
    Zhentao Tan\textsuperscript{\rm 1},
    Ruijie Quan\textsuperscript{\rm 1},
    Yi Yang\textsuperscript{\rm 1}\corresponding
}
\affiliations {
    \textsuperscript{\rm 1}Collaborative Innovation Center of Artificial Intelligence (CCAI), Zhejiang University, China\\
    tanzhentao@zju.edu.cn, quanruijie@zju.edu.cn, yangyics@zju.edu.cn
}

\begin{document}

\maketitle

\begin{abstract}
Neural operators have become a central tool for solving partial differential equations (PDEs), with spectral operators offering efficient global mixing across spatial locations. However, many PDEs contain physics-sensitive local structures that are critical to the underlying physical behavior. For example, in Darcy flow, local material interfaces are often reflected by sharp changes in the permeability field and can strongly influence the solution. Existing spectral operators primarily adapt modal mixing based on center-point representations, making them insufficiently responsive to such localized structural variations. We propose the Edge-Conditioned Spectral Operator (ESO), a novel spectral operator framework that modulates global spectral mixing using local edge-wise variations. By incorporating the Pairwise-Variation Modal Mixer (PVMM) to inject local edge information into spectral mode selection, ESO preserves the global approximation capability of spectral neural operators while enabling the learned kernel to adapt to physics-sensitive local structures. Furthermore, we introduce a task-adaptive Physics-Aware Reweighting (PAR) that emphasizes physically important regions, identified by task-specific physical quantities. Across nine PDE benchmarks, ESO consistently achieves state-of-the-art performance. Visual and region-wise analyses further demonstrate that ESO reduces solution errors near coefficient jumps, high-gradient flow structures, and other physically sensitive regions. The code is available at \url{https://github.com/Tanpig-X/ESO}.
\end{abstract}


\begin{figure}[t]
    \centering
    \includegraphics[width=\linewidth]{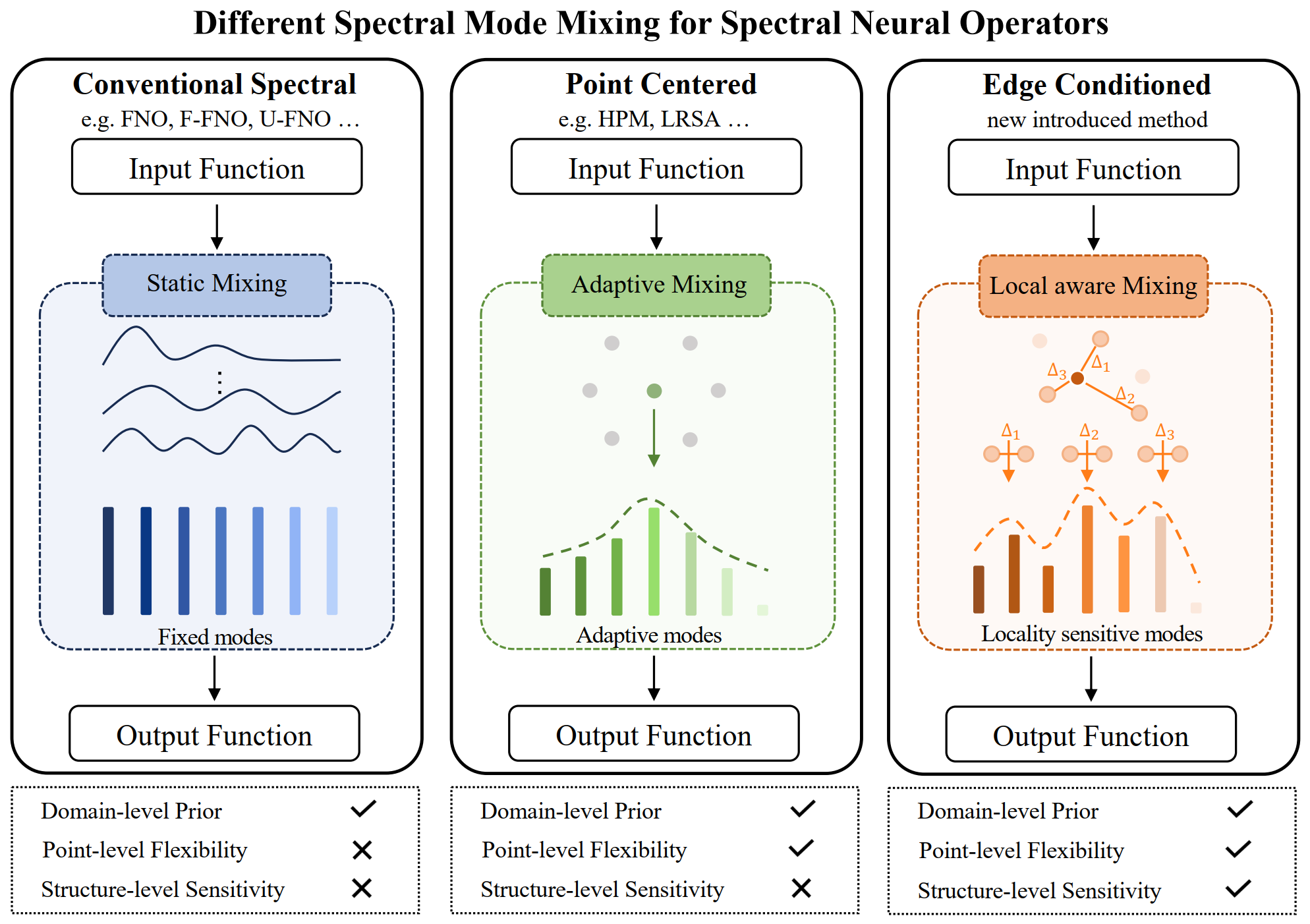}
    \caption{ESO explicitly encodes local pairwise variations for adaptive spectral mixing, enabling the model to capture physics-sensitive local structures more effectively.}
    \label{fig:show}
\end{figure}

\section{Introduction}


Solving partial differential equations (PDEs) is a fundamental research problem with broad real-world applications, ranging from fluid dynamics~\cite{balla2022inverse} to mechanical systems~\cite{haghighat2021physics}. 
Traditional numerical solvers, such as finite element methods~\cite{brenner2008mathematical}, have been widely used to obtain accurate numerical solutions. 
However, these methods often incur huge computational costs for each PDE instance, which limits their practicality in real-time applications. To address this challenge, deep learning has emerged as a promising alternative. In particular, neural operators~\cite{lu2019deeponet,li2020fourier,lu2021learning} provide a data-driven framework for learning continuous mappings between function spaces, enabling efficient approximation of PDE solution operators.

Among neural operators, spectral-based operators have become a prominent family due to their efficient global mixing and strong approximation capability. A representative example is the Fourier Neural Operator (FNO)~\cite{li2020fourier}, which learns PDE solution operators by projecting functions into a truncated Fourier space. Subsequent variants, including PINO~\cite{li2024physics}, Geo-FNO~\cite{li2023fourier}, F-FNO~\cite{tran2021factorized}, GINO~\cite{li2023geometry}, and AM-FNO~\cite{xiao2024amortized}, improve spectral operators from the perspectives of physical regularization, geometric adaptability, or spectral efficiency, leaving the mechanism of spectral mode selection unchanged. More recent attention-based spectral operators, such as AFNO~\cite{guibas2021adaptive}, LSM~\cite{wu2023solving} and HPM~\cite{yue2025holistic} introduce adaptivity by making spectral mode selection depend on point-wise latent features, allowing different spatial locations to emphasize different modes. However, such adaptivity remains primarily point-centered: the modal weights at each location are mainly determined by its own representation. As a result, existing spectral operators insufficiently capture physics-sensitive local structures defined by variations across neighboring points, such as sharp coefficient jumps, leading to performance degradation.

This work introduces the Edge-Conditioned Spectral Operator (ESO), a neural operator that modulates global spectral mixing using local edge-wise variations. The key component of ESO is the Pairwise-Variation Modal Mixer (PVMM), which injects local edge information into spectral mode selection. Specifically, PVMM computes two complementary pairwise statistics that characterize the average local change and the overall variation strength between each point and its neighbors, and uses them to modulate spectral mixing. This design preserves the global approximation capability of spectral neural operators while enabling the learned kernel to adapt to physics-sensitive local structures. Figure~\ref{fig:show} provides an overview of ESO. 
Furthermore, we introduce task-adaptive Physics-Aware Reweighting (PAR), a loss reweighting mechanism that assigns larger weights to errors in physics-sensitive regions, thereby providing explicit supervision for these critical areas. Through extensive experiments across diverse PDE problems, we demonstrate that ESO consistently outperforms state-of-the-art neural operators in overall prediction accuracy, while substantially reducing errors in physics-sensitive regions, as supported by both quantitative results and qualitative visualizations. Overall, our contributions are summarized as follows:
\begin{itemize}
    \item We propose the ESO, which incorporates the PVMM to condition spectral mode selection on local pairwise variations, enabling spectral mixing to adapt to physics-sensitive local structures.
    
    \item We introduce task-adaptive PAR, which explicitly assigns stronger training signals to physics-sensitive regions and complements the implicit edge-aware modulation introduced by PVMM.
    
    \item We conduct extensive experiments on diverse PDE benchmarks, showing that ESO achieves superior overall accuracy compared with state-of-the-art neural operators and significantly reduces errors in physics-sensitive regions.
\end{itemize}

\begin{figure*}[t]
    \centering
    \includegraphics[width=\linewidth]{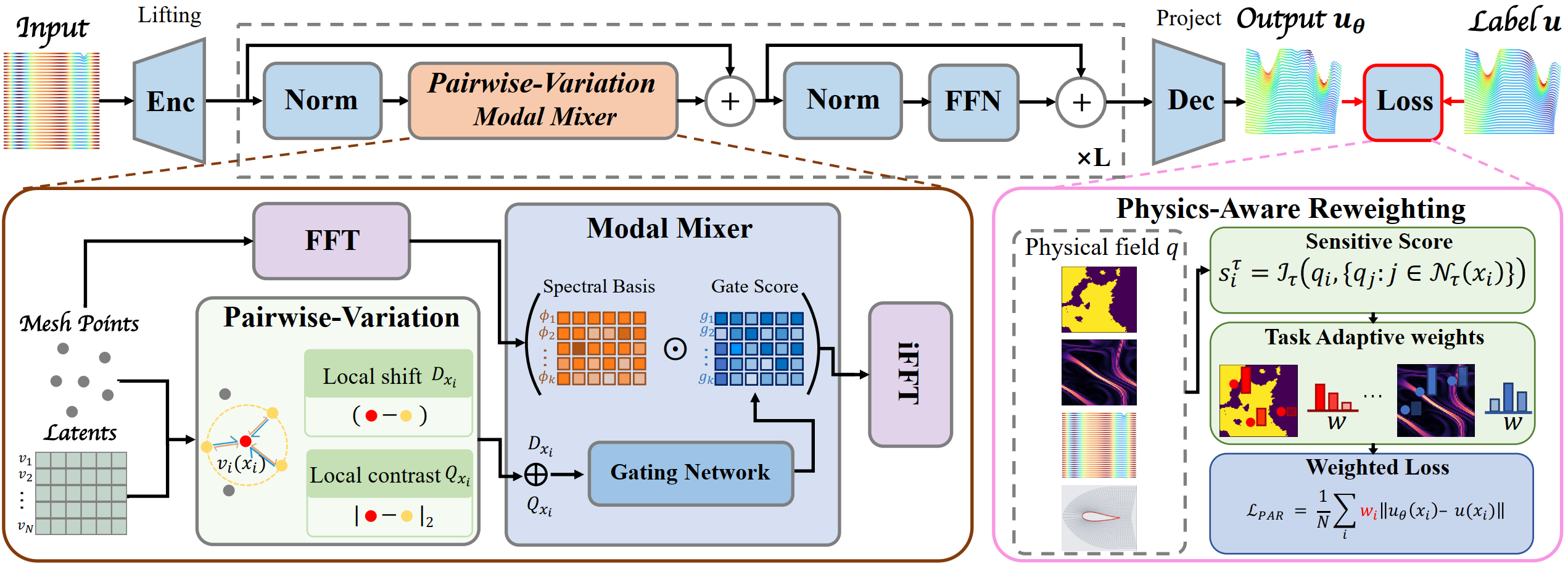}
    \caption{Overview of the Edge-Conditioned Spectral Operator (ESO). ESO computes local pairwise variations between each point and its neighbors, and uses them to modulate spectral mixing through the Pairwise-Variation Modal Mixer (PVMM). The task-adaptive Physics-Aware Reweighting (PAR) further emphasizes physically sensitive regions during training.}
    \label{fig:pipline}
\end{figure*}

\section{Related Work}
Neural operators~\cite{lu2019deeponet,kovachki2023neural} are a fundamental deep learning framework for solving PDEs~\cite{tan2026harnessing}. 
Among them, spectral neural operators model nonlocal interactions in the Fourier domain. 
We review this line of work from the perspective of spectral-mode modeling and explicit mode interactions.



\paragraph{Spectral Neural operators.}  The Fourier Neural Operator (FNO)~\cite{li2020fourier} is a representative spectral neural operator that parameterizes the integral kernel in Fourier space and performs global mixing through truncated spectral modes. 
Following FNO, many variants have been proposed to improve spectral operator learning from different perspectives. 
PINO~\cite{li2024physics} incorporates physics constraints into operator training, while Geo-FNO~\cite{li2023fourier}, GINO~\cite{li2023geometry}, and SFNO~\cite{bonev2023spherical} extend spectral operators to general or spherical geometries. 
Other works improve architectural or spectral efficiency, including F-FNO~\cite{tran2021factorized}, U-FNO~\cite{wen2022u}, MG-TFNO~\cite{kossaifimulti}, and AM-FNO~\cite{xiao2024amortized}. 
Beyond Fourier bases, MWT~\cite{gupta2021multiwavelet}, WNO~\cite{tripura2023wavelet}, LNO~\cite{cao2024laplace} and PILNO~\cite{kim2026physics} explore multiwavelet, wavelet, and Laplace-domain representations for operator learning. 
However, their spectral-mode processing is mostly determined by predefined bases or retained modes. 
Thus, they mainly focus on enhancing the spectral backbone, rather than explicitly modeling input-dependent mode selection conditioned on local physical variations.

\paragraph{Locality and Adaptivity in Spectral Models.}
To improve the flexibility of neural operators, a series of attention-based methods~\cite{hagnberger2024vectorized,wang2024cvit,lee2024inducing} have been proposed to model input-dependent interactions in function spaces. 
Galerkin Transformer~\cite{cao2021choose}, OFormer~\cite{li2022transformer}, GNOT~\cite{hao2023gnot}, FactFormer~\cite{li2023scalable}, ONO~\cite{xiao2023improved}, and Transolver~\cite{wu2024transolver} introduce attention~\cite{vaswani2017attention,kitaev2020reformer} to capture long-range dependencies, handle irregular geometries, or improve operator learning efficiency. More closely related to spectral modeling, recent works incorporate adaptivity into spectral-domain mixing~\cite{hao2024dpot,lee2022fnet,chen2024positional}. 
AFNO~\cite{guibas2021adaptive} performs adaptive token mixing in the Fourier domain. 
LSM~\cite{wu2023solving} introduces latent spectral blocks to model PDE dynamics in a compressed spectral space. 
HPM~\cite{yue2025holistic} further combines spectral representations with point-wise physical states to improve the flexibility of spectral-physical modeling. However, these methods are still mainly point-centered: their modal weights are determined by individual latent features. 
They do not condition spectral mode selection on local neighboring states. 
Recent spectral-attention methods, such as PeSANet~\cite{wan2025pesanet} and SAOT~\cite{zhou2026saot}, enhance spectral operators with adaptive attention or locality-aware designs without modulating spectral basis. 


\section{Method}

To address the limitations of existing spectral neural operators, we propose the Pairwise-Variation Modal Mixer (PVMM), which explicitly computes local edge-wise variations in the latent field and leverages them to modulate spectral mode selection. Furthermore, we introduce a task-adaptive physics-aware reweighting (PAR) that assigns larger weights to physically sensitive regions, further penalizing the critical areas. Finally, we propose Edge-Conditioned Spectral Operators (ESO) and its theoretical analysis.

\paragraph{Problem Setup.} 


Let $\Omega \subset \mathbb{R}^d$ be a bounded open set. Let $a$ and $u$ denote the input and output functions, respectively, belonging to the Banach spaces $\mathcal{A}(\Omega; \mathbb{R}^{d_a})$ and $\mathcal{U}(\Omega; \mathbb{R}^{d_u})$, i.e., $a \in \mathcal{A}(\Omega; \mathbb{R}^{d_a})$ and $u \in \mathcal{U}(\Omega; \mathbb{R}^{d_u})$. Suppose that $\mathcal{G}$ is an operator mapping inputs to outputs, $\mathcal{G}: \mathcal{A} \rightarrow \mathcal{U}, a \mapsto u$. In practice, it is intractable to represent functions exactly over a continuous domain. Therefore, we represent $\Omega$ by a discretized set $\tilde{\Omega} = \{x_i\}_{i=1}^{N}$, which may correspond to a structured grid or an unstructured mesh. The learning task is to approximate the operator $\mathcal{G}$ with a parameterized neural network $\mathcal{G}_{\theta}$, where $\theta$ denotes the trainable parameters.

\paragraph{Preliminary: Spectral Mixing.} Given a latent field function $v_t \in \mathbb{R}^{N \times d_v}$ at layer $t$ on points $\{x_i\}_{i=1}^{N}$, FNO-like spectral methods~\cite{li2020fourier,kovachki2023neural,chen2024learning} compute the output field function $v_{t+1}$ through:
\begin{equation}                                                                                                                                                                                                                                                                                                     
      v_{t+1}(x_i)                                                                                                                                                                                                                                                                                                     
      =                                                                                                                                                                                                                                                                                                                
      \sigma\left(                                                                                                                                                                                                                                                                                                     
      Wv_t(x_i)                                                                                                                                                                                                                                                                                                        
      +                                                                                                                                                                                                                                                                                                                
      (\mathcal{S}v_t)(x_i)                                                                                                                                                                                                                                                                                   
      \right),                                                                                                                                                                                                                                                                                                         
  \end{equation}    
where $W$ is a learnable parameter, and $\mathcal{S}$ denotes spectral mixing. The spectral mixing term is computed as follows:

\begin{equation}       
\label{spectral mixing}
\begin{split}
    (\mathcal{S}v_t)(x_i)
      &=
      \sum_{k\in\mathcal{M}_K}
      \phi_k(x_i) R_k \widehat{v}_t(k), \\
      \widehat{v}_t(k)
      &=
      \sum_{j=1}^{N}
      \phi_k(x_j)v_t(x_j).    
\end{split}
\end{equation}
  where $\mathcal{M}_K$ is the truncated set of retained $K$ spectral modes,
  $\phi_k \in \mathbb{C}^{N}$ is a Fourier basis function, and $R_k \in \mathbb{C}^{d_v \times d_v}$ is a learnable mode-wise parameter. In adaptive spectral variants~\cite{yue2025holistic}, the basis is modulated by a point-centered score $\pi$ with a learnable mapping $f_\theta$,                                                                                                               
  \begin{equation}                                                                                                                                                                                                                                                                                                     
      \pi_{i,k}                                                                                                                                                                                                                                                                                                          
      =                                                                                                                                                                                                                                                                                                                
      \operatorname{softmax}_k                                                                                                                                                                                                                                                                                         
      \left(
      f_{\theta}(v_t(x_i))
      \right),
  \end{equation}
and calibrates the adaptive spectral basis as
  \begin{equation}
      \widetilde{\phi}_{i,k}
      =
      \pi_{i,k}\phi_k(x_i).
  \end{equation}
  The adaptive basis is then fused in equation \ref{spectral mixing} for further spectral mixing. However, such point-centered conditioning overlooks local variations, e.g., sharp permeability contrasts in Darcy flow, leading to degraded solution accuracy.

\begin{table*}[t]
    \centering
    {\fontsize{8.8pt}{9pt}\selectfont
    \begin{tabular}{cccccc}
    \toprule
         \textbf{Benchmarks}& \textbf{Input}& \textbf{Output}& \textbf{Geometry}& \textbf{Resolution}& \textbf{Temporal} \\
    \hline
         Darcy Flow& Permeability coefficient & Pressure & Structured meshes (regular) & 85 $\times$ 85& - \\
         Navier-Stokes& Historical vorticity & Future vorticity & Structured meshes (regular) & 64 $\times$ 64& 10 \\
         Airfoil& Airfoil meshes & Mach number & Structured meshes (irregular) & 221 $\times$ 51& - \\
         Plasticity& Rigid die meshes& Plastic deformation& Structured meshes (irregular) & 101 $\times$ 31& 20 \\
         Irregular Darcy& Permeability coefficient& Pressure& Unstructured meshes & 2290 Nodes& - \\
         Pipe Turbulence& Previous velocity& Next-step velocity& Unstructured meshes & 2673 Nodes& 1 \\
         Heat Transfer& Initial / boundary temperature& Future 3D temperature& Unstructured meshes & 7199 Nodes& 1 \\
         Composite& Thermal loading field& Deformation field& Unstructured meshes & 8232 Nodes& - \\
         Blood Flow & Inlet velocity and outlet pressure& Blood velocity& Unstructured meshes & 1656 Nodes& 121 \\
    \bottomrule
    \end{tabular}}
    \caption{Summary of the nine PDE benchmarks, including input-output variables, mesh geometry, spatial resolution, and temporal prediction dimension. The benchmark covers four structured-mesh problems and five unstructured-mesh problems.}
    \label{tab:benchmarks}
\end{table*}

\paragraph{Pairwise-Variation Modal Mixer}\hfill\par
\noindent
To address the limitation, we propose local edge-aware modulation via the Pairwise-Variation Modal Mixer (PVMM): 
$\mathcal{E}[v_t]: \mathbb{R}^{N \times d_v} \rightarrow \mathbb{R}^{N \times K}$. 
For each point $x_i$ and its corresponding latent feature $v_t(x_i)$ at layer $t$, PVMM constructs local shift statistic $D_{x_i}$ and local contrast statistic $Q_{x_i}$:
\begin{equation}
    D_{x_i} =
    \sum_{x_j \in \mathcal{N}(x_i)}
    \frac{1}{|\mathcal{N}(x_i)|}
    ( v_t(x_j) - v_t(x_i) ),
\end{equation}
\begin{equation}
    Q_{x_i} =
    \sum_{x_j \in \mathcal{N}(x_i)}
    \frac{1}{|\mathcal{N}(x_i)|}
    (v_t(x_j) - v_t(x_i))
    \odot
    ( v_t(x_j) - v_t(x_i)),
\end{equation}
where $\mathcal{N}(x_i)$ denotes the neighborhood of point $x_i$, and $\odot$ denotes element-wise multiplication. 
$D_{x_i}$ measures the signed average difference between the center state and its neighbors, preserving directional information for asymmetric changes near interfaces. 
However, signed differences may cancel out during averaging. 
To address this, $Q_{x_i}$ measures the average squared magnitude of local differences, highlighting locations where the center state differs strongly from its surroundings regardless of direction, such as material jumps in Darcy flow or vortical structures in fluid dynamics.

Based on these two statistics, the PVMM at point $x_i$ is computed as
\begin{equation}
    \mathcal{E}[v_t](x_i)
    =
    \psi_{\theta_E}
    \left(
    \operatorname{concat}
    \left(
    v_t(x_i), D_{x_i}, Q_{x_i}
    \right)
    \right),
\end{equation}
where $\psi_{\theta_E}: \mathbb{R}^{3d_v} \rightarrow \mathbb{R}^{K}$ is an gating network that maps local variation statistics to spectral modal logits. 
With PVMM, the final edge-conditioned spectral score $\pi_{i,k}^E$ is then defined as
\begin{equation}
    g^E_i
    =
    \lambda_E f_{\theta}(v_t(x_i))
    +
    (1-\lambda_E)\mathcal{E}[v_t](x_i),
\end{equation}
\begin{equation}
    \pi^E_{i,k}
    =
    \operatorname{softmax}_{k}(g^E_i),
\end{equation}
where $\lambda_E \in [0,1]$ balances the point-conditioned spectral component and the edge-conditioned component. 
Finally, the edge-conditioned spectral basis is obtained as:
\begin{equation}
    \widetilde{\Phi}_{i,k}^E
    =
    \pi^E_{i,k}\phi_k(x_i),
\end{equation}
or equivalently,
\begin{equation}
\label{PVMM}
    \widetilde{\Phi}^E
    =
    \Pi^E \odot \Phi,
\end{equation}
where $\Phi_{i,k}^E=\phi_k^E(x_i)$ and 
$\Pi^E=\{\pi^E_{i,k}\}_{i=1,k=1}^{N,K}$. 
With this edge-conditioned spectral basis, we can better capture abrupt changes in sensitive regions and alleviate the over-smoothing effect commonly observed in conventional spectral methods~\cite{li2020fourier}. Following~\cite{yue2025holistic}, we define $\Phi$ using Laplace--Beltrami eigenfunctions, which provide a flexible spectral representation across structured and unstructured mesh problems.

\paragraph{Task-adaptive Physics-Aware Reweighting}\hfill\par
\noindent
To further enforce accuracy, we introduce a task-adaptive Physics-Aware Reweighting (PAR), which explicitly penalizes prediction errors in physics-sensitive regions.

PAR builds a training weight map from physical fields that are already available in each benchmark. Let $q_i$ be such a task-available field at point $x_i$. For example, in Darcy flow, $q_i=a_i$ is the input permeability coefficient. PAR first convert $q$ into a nonnegative sensitivity score:
\begin{equation}
    s_i^\tau
    =
    \mathcal{I}_{\tau}
    (
    q_i,\{q_j:j\in\mathcal{N}_{\tau}(x_i)\}
    ),
    \label{eq:general-indicator}
\end{equation}
where $\tau$ denotes the task type, $\mathcal{N}_{\tau}(x_i)$ denotes the task-specific local neighborhood of $x_i$, and $\mathcal{I}_{\tau}$ is a nonnegative local sensitivity functional. In our implementation, we instantiate $\mathcal{I}_{\tau}$ as the $\ell_1$ norm, i.e., $\mathcal{I}_{\tau}=|\cdot|_1$. We set $\mathcal{N}_{\tau}=\mathcal{N}_{\mathrm{str}}$ for structured-mesh problems and $\mathcal{N}_{\tau}=\mathcal{N}_{\mathrm{unstr}}$ for unstructured-mesh problems. The task-specific choice of $q$ is provided in Appendix.

Given $s_i^\tau$, PAR constructs normalized weights by
\begin{equation}
    \bar{s}^\tau
    =
    \frac{1}{N}
    \sum_{i=1}^{N}
    s_i^\tau,
\end{equation}
\begin{equation}
    \widetilde{w}_i
    =
    1+
    \frac{s_i^\tau}{\bar{s}^{\tau}+\varepsilon},
    w_i
    =
    \frac{\widetilde{w}_i}
    {\frac{1}{N}\sum_{j=1}^{N}\widetilde{w}_j}.
\end{equation}
The unnormalized weight $\widetilde{w}_i$ increases monotonically with $s_i^\tau$, while the division by $\bar{s}^\tau$ makes the weight depend on relative sensitivity within each sample. The PAR objective is then written as:
\begin{equation}
    \mathcal{L}_{\mathrm{PAR}}(\theta)
    =
    \frac{1}{N}
    \sum_{i=1}^{N}
    w_i
    \|u_{\theta}(x_i)-u(x_i)\|_2^2.
\end{equation}
Unlike PVMM, which implicitly encodes physics-sensitive structures through feature modulation, PAR explicitly allocates stronger training signals to physically important regions, thereby further improving solution accuracy.

\paragraph{Edge-Conditioned Spectral Operator}\hfill\par
\noindent
The Edge-Conditioned Spectral Operator (ESO) model is generally based on spectral methods, with conventional spectral mixing replaced by the proposed PVMM.

\begin{table*}[t]
    \centering
    \begin{tabularx}{\textwidth}{C{0.25\textwidth}|YYYY}
    \toprule
    \textbf{Model}& \textbf{Darcy Flow}&\textbf{Airfoil}&\textbf{Navier-Stokes}&\textbf{Plasticity}  \\
    \hline
    Geo-FNO (JMLP 2023)& 1.08e-2&1.38e-2&1.56e-1&7.40e-3  \\
    GINO (NIPS 2023) & 8.23e-3&6.08e-3&1.29e-1&3.94e-3  \\
    GNOT (ICML 2023)& 1.05e-2&7.60e-3&1.38e-1&-\\
    RIGNO (NIPS 2025)& 8.01e-3&5.72e-3&1.04e-1&3.25e-3 \\
    \hline
    Galerkin (NIPS 2021)& 8.40e-3&1.18e-2&1.40e-1&1.20e-2  \\
    Oformer (TMLR 2023) & 1.24e-2&1.83e-2&1.71e-1&1.70e-3  \\
    ONO (NIPS 2024)& 7.60e-3&6.10e-3&1.20e-1&4.80e-3  \\
    Transolver (ICML 2024)& 5.70e-3&5.30e-3&9.00e-2&1.23e-3 \\
    \hline
    FNO (ICLR 2021)& 1.08e-2&-&1.56e-1&-  \\
    F-FNO (ICLR 2023) & 7.70e-3&7.80e-3&2.32e-1&4.39e-3  \\
    LSM (ICML 2023)& 6.50e-3&5.90e-3&1.54e-1&2.50e-3  \\
    NORM (NSO 2024)& 9.71e-3&5.44e-3&1.15e-1&4.39e-3 \\
    \hline
    HPM (ICML 2025)& 4.59e-3&4.72e-3&7.34e-2&8.00e-4  \\
    SAOT (AAAI 2026)& 4.90e-3&4.80e-3&6.88e-2& 8.21e-4  \\
    LRSA (ICML 2026)& \underline{4.30e-3}&\underline{4.20e-3}&\underline{4.84e-2}& \underline{5.67e-4}  \\
    \hline
    \rule[-0.8ex]{0pt}{3.2ex}ESO& \textbf{3.52e-3}&\textbf{3.93e-3}&\textbf{4.77e-2}&\textbf{4.90e-4}  \\
    \bottomrule
    \end{tabularx}
    \caption{Comparison on structured mesh problems. The best result is \textbf{bold}, and the second result is \underline{underline}.}
    \label{tab:structure_exp}
\end{table*}

\paragraph{Overall Architecture.} 

Following previous works~\cite{li2020fourier,yue2025holistic}, we implement ESO using a standard encoder--processor--decoder architecture, where PVMM modules serve as the intermediate processor blocks, and projection layers $E$ and $D$ are placed at the beginning and the end, respectively:
\begin{equation}
    \mathcal{G}_\theta^{ESO} = D \circ M^{PVMM}_l \circ ... \circ M^{PVMM}_2 \circ M^{PVMM}_1 \circ E,
\end{equation}
where $M^{PVMM}_t$ can be written in matrix formulation as :
\begin{equation}
\begin{split}
    \mathbf{V}_{t+1}
    &=
    \sigma(
    \mathbf{V}_t \mathbf{W}
    + \mathcal{S}^E \mathbf{V}_t
    ), \\
        \mathcal{S}^E \mathbf{V}_t &= \widetilde{\Phi}^E
    \mathbf{R}
    (
    (\widetilde{\Phi}^E)^\top \mathbf{V}_t
    ).
\end{split}
\end{equation}
$\mathbf{W} \in \mathbb{R}^{d_v \times d_v}$ and $\mathbf{R} \in \mathbb{C}^{K \times d_v \times d_v}$ are learnable parameters and $\widetilde{\Phi}^E \in \mathbb{C}^{N \times K} $ refers to equation \ref{PVMM}, the pipeline is shown in Figure~\ref{fig:pipline}.

\paragraph{Calculation of $\mathcal{N}(x_i)$.} For structured meshes, every node has a logical grid index even when its physical coordinates are not uniformly spaced. If $x_i$ corresponds to grid coordinate $(p_i,q_i)$, we define $\mathcal{N}_{\mathrm{str}}(x_i)$ as follows:
\begin{equation}
    \mathcal{N}_{\mathrm{str}}(x_i)
    =
    \{
    (p_i+\Delta p,\ q_i+\Delta q)\in\tilde{\Omega}:
    \Delta p,\Delta q\in\{-1,0,1\}
    \},
\label{eq:structured-neighborhood}
\end{equation}
in implementation, this neighborhood aggregation is computed by grouped convolution~\cite{krizhevsky2012imagenet} with a $3\times3$ kernel whose center entry is zero.
\begin{figure}[t]
    \centering
    \resizebox{0.90\linewidth}{0.2\height}{
        \includegraphics{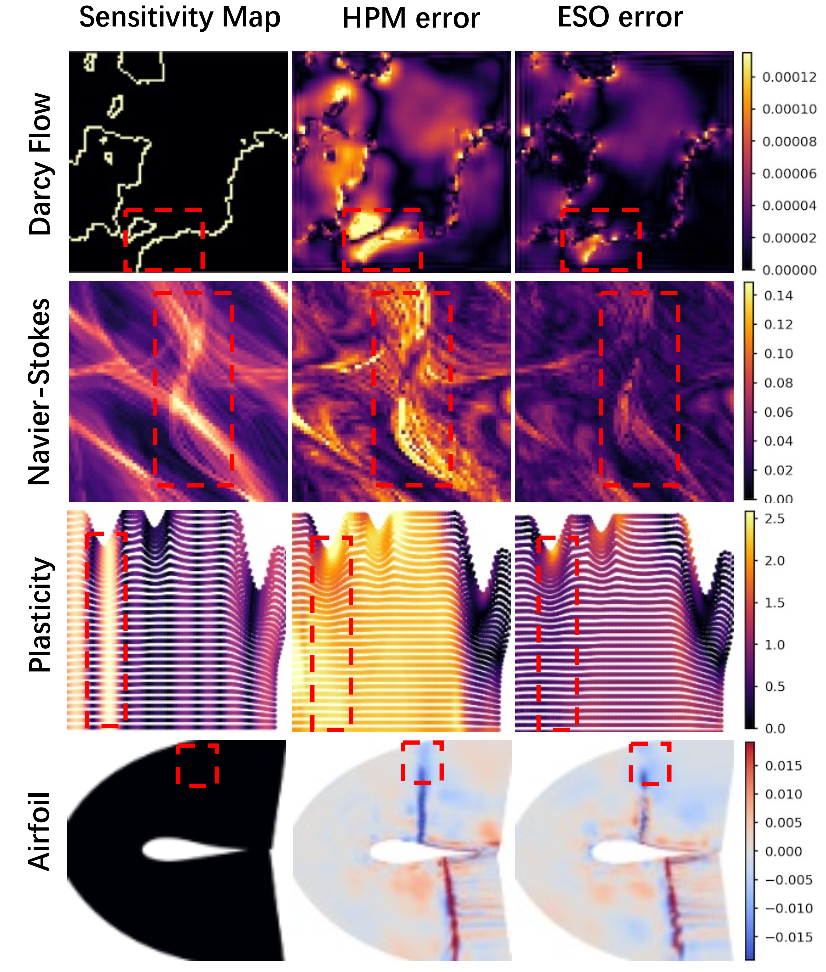}
    }
    \caption{Qualitative comparison on structured PDE benchmarks.
  The sensitivity map highlights task-specific local structures, with sensitive regions' examples outlined in red.
  ESO reduces errors in physics-sensitive regions. The brighter the color, the larger the error.}
    \label{fig:error_map}
\end{figure}

For unstructured meshes, nodes do not have regular grid indices. We first build a fixed neighbor graph by k-nearest neighbors (kNN)~\cite{belkin2003laplacian} in a precomputed mesh spectral-coordinate space. Specifically, we use the first $d_z$ Laplace--Beltrami eigenfunctions~\cite{chen2024learning} as node coordinates:
\begin{equation}
    Z^{\mathrm{mesh}}
    =
    [\phi_1,\ldots,\phi_{d_z}]
    \in\mathbb{R}^{N\times d_z},
    z_i = Z^{\mathrm{mesh}}_{i,:}\in\mathbb{R}^{d_z},
    \label{eq:mesh-spectral-coordinate}
\end{equation}
where $z_i$ is the row vector collecting the eigenvector values at node $x_i$. Thus $z_i$ is a fixed geometric coordinate derived from the mesh connectivity and node geometry.We thus define
\begin{equation}
    \mathcal{N}_{\mathrm{unstr}}(x_i)
    =
    \operatorname{kNN}_{K_n}
    (
    z_i;\{z_j\}_{j=1}^{N}
    ),
    \label{eq:unstructured-neighborhood}
\end{equation}
where $K_n$ is the number of neighbors. Unless otherwise stated, we set $k_n=8$ in all experiments.


\paragraph{Theoretical Perspective}\hfill\par
\noindent
We first show that ESO is at least as expressive as conventional spectral-based methods.


\begin{Theorem}[No-worse approximation]
\label{thm:better-approx}
Let $\mathcal{F}_{\mathrm{base}}$ denote the conventional spectral-based operators, and let $\mathcal{F}_{\mathrm{edge}}$ denote the edge-conditioned operators. Then, for any data distribution $\mathcal{D}$ and any nonnegative loss function $\ell$, we have
\begin{equation}
    \inf_{f \in \mathcal{F}_{\mathrm{edge}}}
    \mathbb{E}_{(x,y)\sim \mathcal{D}}
    [
        \ell(f(x),y)
    ]
    \leq
    \inf_{f \in \mathcal{F}_{\mathrm{base}}}
    \mathbb{E}_{(x,y)\sim \mathcal{D}}
    [
        \ell(f(x),y)
    ].
\end{equation}
\end{Theorem}

\begin{table*}[t]
    \centering
    \renewcommand{\arraystretch}{1.12}
    {\fontsize{8.5pt}{10pt}\selectfont
    \begin{tabularx}{\textwidth}{C{0.24\textwidth}|YYYYY}
    \toprule
    \textbf{Model} & 
    \textbf{\mbox{Irregular Darcy}} &
    \textbf{\mbox{Pipe Turbulence}} &
    \textbf{\mbox{Heat Transfer}} &
    \textbf{Composite} & 
    \textbf{\mbox{Blood Flow}} \\
    \hline
    GraphSAGE (NIPS 2017) & 6.73e-2 & 2.36e-1 & - & 2.09e-1 & -  \\
    DeepOnet (Arxiv 2019) & 1.36e-2 & 9.36e-2 & 7.20e-4 & 1.88e-2 & 8.93e-1  \\
    \mbox{POD-DeepOnet (CMAME 2022)} & 1.30e-2 & 2.59e-2 & 5.70e-4 & 1.44e-2 & 3.74e-1  \\
    FNO (ICLR 2021) & 3.83e-2 & 3.80e-2 & - & - & -  \\
    NORM (NSO 2024) & 1.05e-2 & 1.01e-2 & 2.70e-4 & 9.99e-3 & 4.82e-2  \\
    HPM (ICML 2025) & \underline{7.38e-3} & 8.26e-3 & 1.84e-4 & 9.34e-3 & \underline{2.89e-2}  \\
    LRSA (ICML 2026) & 7.56e-3 & \underline{4.89e-3} & \underline{1.49e-4} & \underline{8.71e-3} & - \\
    \hline
    \rule[-0.8ex]{0pt}{3.2ex}ESO & \textbf{6.71e-3} & \textbf{4.32e-3} & \textbf{9.75e-5} & \textbf{7.93e-3} & \textbf{2.76e-2} \\
    \bottomrule
    \end{tabularx}}
    \caption{Comparison on unstructured mesh problems. The best result is \textbf{bold}, and the second-best result is \underline{underlined}.}
    \label{tab:unstructured_exp}
\end{table*}

Theorem~\ref{thm:better-approx} shows that, under the hypothesis-class inclusion
assumption, introducing edge-conditioned modulation does not
reduce the best achievable approximation capability of the
base spectral operator.

\begin{Theorem}[Modal-gating separation]
\label{thm:separation}
Suppose there exist two admissible inputs $a$ and $\widetilde{a}$
and a node $x_i$ such that
\begin{equation}
    h_i(a)=h_i(\widetilde{a}),
    \qquad
    e_i(a)\neq e_i(\widetilde{a}).
\end{equation}
Any point-conditioned modal gate
$\pi_i^{\mathrm{P}}(a)$
must satisfy $\pi_i^{\mathrm{P}}(a)
    =
    \pi_i^{\mathrm{P}}(\widetilde{a}).$ If a target modal-weight mapping satisfies
\begin{equation}
    \left\|
    \pi_i^\star(a)-\pi_i^\star(\widetilde{a})
    \right\|_2
    \geq \Delta_\pi,
\end{equation}
then any point-conditioned gate incurs an error of at least
$\Delta_\pi/2$ on one of the two inputs. In contrast, a
sufficiently expressive edge-conditioned gate depending on both
$h_i$ and $e_i$ can assign distinct modal weights to them.
\end{Theorem}

\begin{table}[t]
\centering
\begin{tabular*}{\linewidth}{@{\extracolsep{\fill}}l|cc}
\toprule
Method & Darcy Flow & Navier--Stokes  \\
\midrule
Baseline & 4.59e-3 & 7.34e-2  \\
+ PAR only & 4.32e-3 & 6.91e-2  \\
+ PVMM only & 3.78e-3 & 5.06e-2  \\
ESO: PVMM + PAR & \textbf{3.52e-3} & \textbf{4.77e-2}  \\
\bottomrule
\end{tabular*}
\caption{Ablation study of PVMM and PAR components on Darcy Flow and Navier--Stokes. The full ESO model combining PVMM and PAR achieves the lowest relative $L_2$ error.}
\label{tab:ablation}
\end{table}

Theorem~\ref{thm:separation} characterizes a limitation of
point-conditioned modal gating. Although spectral mixing aggregates global
information, its point-conditioned gate cannot respond to
different neighboring structures when the center representation
is unchanged. ESO can remove this invariance by conditioning modal
selection jointly on $h_i$ and $e_i$.

\section{Experiments}
In this section, we first introduce the experimental setups. We then compare ESO with several baseline methods across nine PDE benchmarks with structured and unstructured mesh to demonstrate its effectiveness. Finally, we conduct ablation studies to validate each design component.

\paragraph{Experimental Setups}\hfill\par
\noindent
\paragraph{Benchmarks.} We evaluate ESO on nine standard PDE benchmarks spanning from structured meshes problems with regular grids: Darcy Flow, Navier-Stokes and with irregular grids: Airfoil, Plasticity to unstructured meshes problems: Irregular Darcy, Pipe Turbulence, Heat Transfer, Composite, Blood Flow following the protocol in~\cite{yue2025holistic}. These tasks are controlled by different PDE from steady-state problems to time-series problems. A summary of the benchmarks are in Table \ref{tab:benchmarks}, more descriptions can be found in the Appendix.

\paragraph{Metrics.} Unless otherwise specified, we report the test relative $L_2$ error, defined as $|\hat{\mathbf{u}}-\mathbf{u}|_2 / |\mathbf{u}|_2$. To ensure a fair comparison, all baselines are matched with ESO under comparable parameter budgets and consistent training configurations. More experimental details are provided in Appendix.

\paragraph{Structured Mesh Problems}\hfill\par
\noindent
We first evaluate ESO against existing neural operators on structured-mesh problems, where the meshes are aligned with standard rectangular grids.

\paragraph{Baselines.} We compare ESO against four categories of baselines. \textbf{Conventional spectral operators} including FNO~\cite{li2020fourier}, F-FNO~\cite{tran2021factorized}, LSM~\cite{wu2023solving}, and NORM~\cite{chen2024learning}, which primarily improve the backbones with fixed basis. \textbf{Adaptive spectral operators} including HPM~\cite{yue2025holistic}, SAOT~\cite{zhou2026saot}, and LRSA~\cite{yang2026simple}, which perform spectral modulation or local adaptation. \textbf{Graph-based operators} including Geo-FNO~\cite{li2023fourier}, GINO~\cite{li2023geometry}, GNOT~\cite{hao2023gnot}, and RIGNO~\cite{mousavi2026rigno}, which propagate information over graph structures in physical fields.
\textbf{Attention-based operators} including Galerkin~\cite{cao2021choose}, Oformer~\cite{li2022transformer}, ONO~\cite{xiao2023improved}, and Transolver~\cite{wu2024transolver} which capture long-range dependencies through attention mechanisms.
\begin{table}[t]
\centering
\begin{tabular*}{\linewidth}{@{\extracolsep{\fill}}l|cc}
\toprule
PVMM Variant & Darcy Flow & Navier--Stokes  \\
\midrule
Points only & 4.59e-3 & 7.34e-2  \\
$D_{x_i}$ only & 4.16e-3 & 5.43e-2  \\
$Q_{x_i}$ only & 3.92e-3 & 5.25e-2  \\
$D_{x_i} + Q_{x_i}$ & \textbf{3.78e-3} & \textbf{5.06e-2} \\
\bottomrule
\end{tabular*}
\caption{Ablation study of PVMM variants on Darcy Flow and Navier--Stokes. The full variant combining $D_{x_i}$ and $Q_{x_i}$ achieves the lowest relative $L_2$ error.}
\label{tab:pvmm_variant}
\end{table}

\paragraph{Quantitative results.} Table~\ref{tab:structure_exp} reports the quantitative results. ESO achieves the best performance across all tasks. Conventional spectral operators provide efficient global modeling but are limited by their relatively fixed spectral representations, leading to weaker performance than adaptive spectral methods. Graph-based operators and attention-based operators can model irregular connectivity or long-range dependencies, but they do not explicitly adapt spectral mixing to local physical variations, which limits their performance. Among all baselines, adaptive spectral operators are the most competitive, with LRSA obtaining the second-best results on all four benchmarks. ESO further improves over LRSA by introducing edge-conditioned spectral modulation, which enables the model to incorporate local variations into spectral-mode selection while preserving efficient global mixing.

\paragraph{Qualitative results.} Figure~\ref{fig:error_map} visualizes the relation between task-specific sensitivity maps and prediction errors, where representative local physics-sensitive regions are outlined in red for clarity. The benefit of ESO is most evident on Darcy Flow and Navier--Stokes, where the error patterns are well aligned with the highlighted sensitive structures. In Darcy Flow, HPM produces large errors near sharp permeability interfaces, while ESO reduces errors around these coefficient-jump regions. In Navier--Stokes, high-sensitivity regions correspond to vortical and shear-dominated structures, where ESO also yields visibly lower errors than HPM. For Plasticity, the sensitive regions are deformation-localization patterns induced by the die geometry; ESO reduces displacement-magnitude errors over the deformed body, suggesting that edge-conditioned mixing is also useful for geometry-induced material responses. Airfoil is a smoother geometry-dominated case, where local sensitivity is less sharply defined by jump indicators, but ESO still moderately reduces errors near the airfoil.

\paragraph{Unstructured Mesh Problems}\hfill\par
\noindent
We then evaluate ESO on unstructured-mesh problems, where the meshes are irregular triangle meshes.

\paragraph{Baselines.} We compare ESO against seven baselines, including GraphSAGE~\cite{hamilton2017inductive}, DeepOnet~\cite{lu2019deeponet}, POD-DeepOnet~\cite{lu2022comprehensive}, FNO~\cite{li2020fourier}, NORM~\cite{chen2024learning}, HPM~\cite{yue2025holistic}, LRSA~\cite{yang2026simple}.
\begin{figure}[t]
    \centering
    \begin{subfigure}{0.48\linewidth}
        \centering
        \scalebox{1}[1.4]{\includegraphics[width=\linewidth]{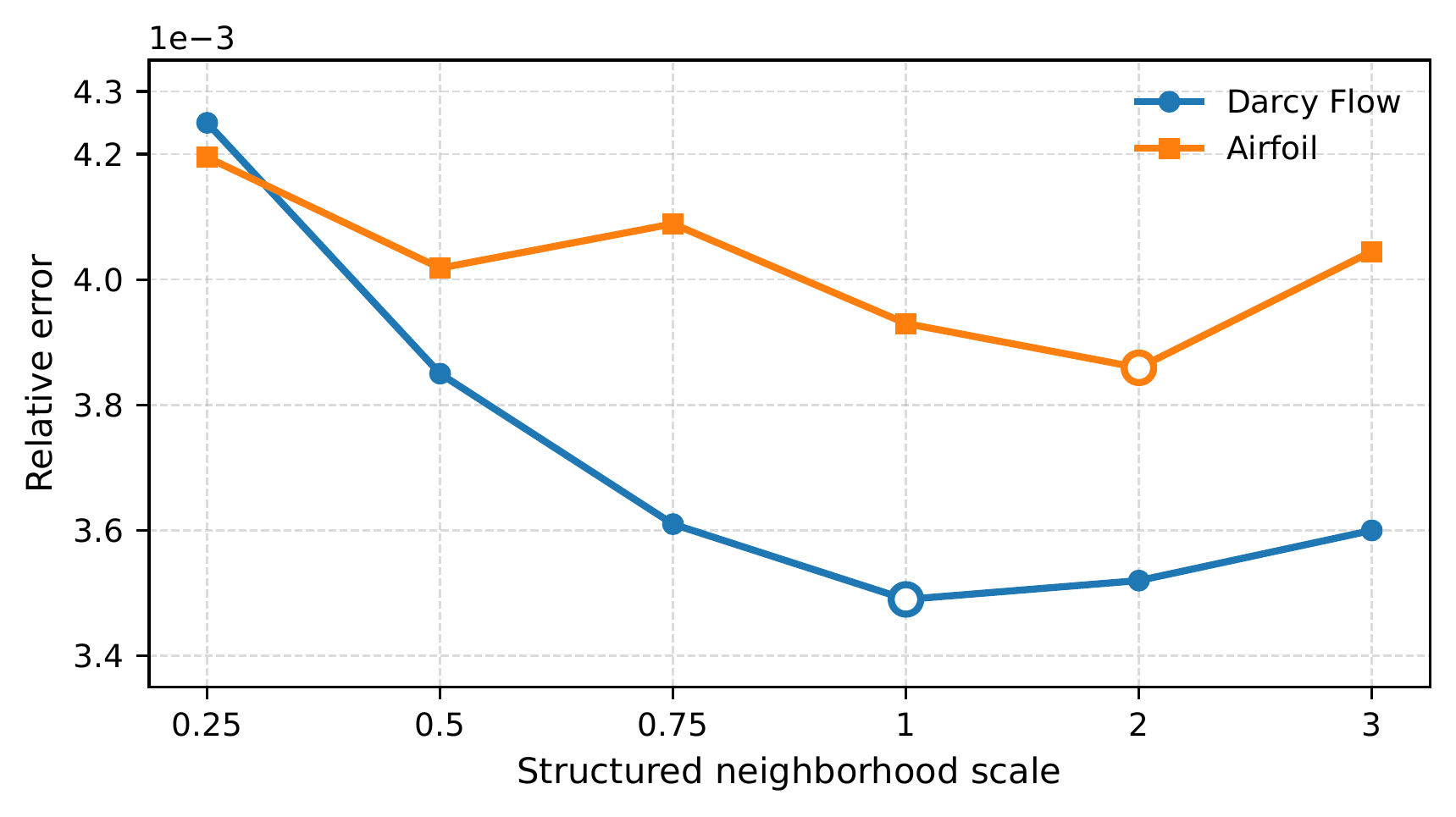}}
        \caption{Structured neighbor scale}
        \label{fig:structured_scale_curve}
    \end{subfigure}
    \begin{subfigure}{0.48\linewidth}
        \centering
        \scalebox{1}[1.4]{\includegraphics[width=\linewidth]{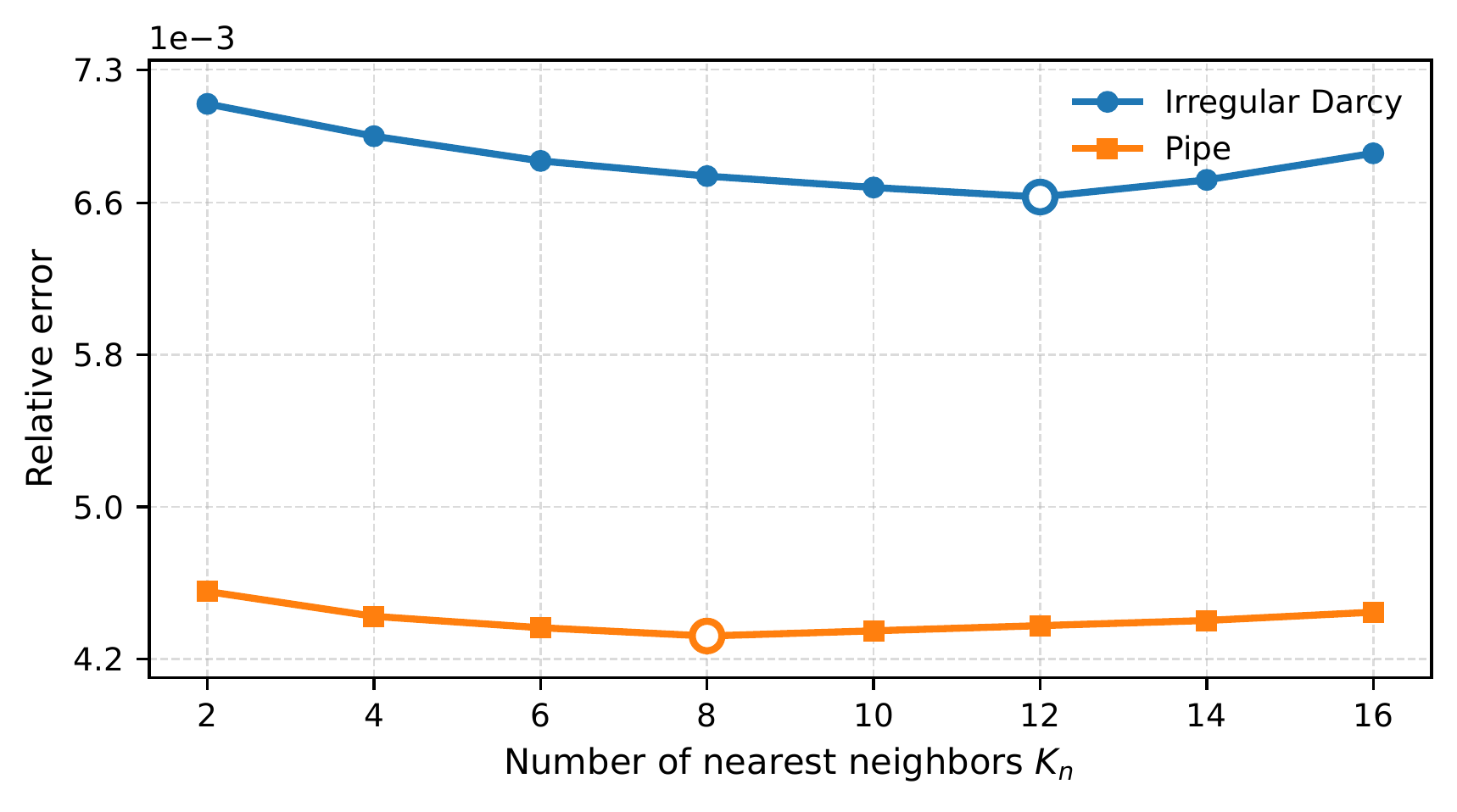}}
        \caption{Unstructured neighbor scale}
        \label{fig:kn_scale_curve}
    \end{subfigure}
    \hfill
    \caption{Ablation study on neighborhood scale choices for unstructured and structured meshes. 
The results show that properly selected local neighborhoods help capture physics-sensitive variations and improve model performance.}
    \label{fig:scale_curve}
\end{figure}
\paragraph{Results.} As shown in Table~\ref{tab:unstructured_exp}, ESO achieves the best results on all five unstructured-mesh benchmarks. Traditional graph-based and operator-learning baselines perform relatively worse, while geometry-aware and adaptive spectral methods such as NORM, HPM, and LRSA are more competitive. ESO further improves over the best baseline on every task, with especially large gains on Heat Transfer. These results demonstrate that the proposed edge-conditioned spectral modulation can effectively handle local geometric variations and complex physical patterns on unstructured meshes.

\paragraph{Ablation Studies}\hfill\par
\noindent
In this section, we evaluate each designs component of ESO, including the PVMM and the PAR. We conduct ablation studies on both structured and unstructured mesh problems.
\paragraph{Main Components.} Both PVMM and PAR improve the baseline on Darcy Flow and Navier--Stokes. As shown in Table~\ref{tab:ablation}, PVMM yields larger gains than PAR, indicating that edge-conditioned spectral modulation is the main contributor to performance improvement. Meanwhile, PAR further reduces the error by emphasizing physics-sensitive regions during training. The full ESO model, which combines PVMM and PAR, achieves the best results on both benchmarks, confirming the complementary effects of the two components.

\paragraph{Ablation of PVMM Variants.}
As shown in Table~\ref{tab:pvmm_variant}, the point-only variant performs the worst, indicating that point-wise features alone are insufficient for adaptive spectral modulation. 
Adding $D_{x_i}$ improves performance by introducing first-order local differences, while $Q_{x_i}$ further captures variation magnitude and mitigates sign cancellation. 
The full variant combining $D_{x_i}$ and $Q_{x_i}$ achieves the lowest errors on both Darcy Flow and Navier--Stokes, demonstrating that signed differences and quadratic local variations provide complementary cues for spectral modulation.
\paragraph{Ablation of Neighborhood Scales.} Neighborhood scale determines the local pairwise variation available to PVMM. 
As shown in Fig.~\ref{fig:structured_scale_curve}, Darcy Flow performs best with a complete first-order neighborhood, while smaller or larger stencils either miss permeability jumps or introduce irrelevant regions. 
Airfoil is less sensitive due to its smoother geometry. 
For unstructured meshes in Fig.~\ref{fig:kn_scale_curve}, the optimal $K_n$ is task-dependent, with Irregular Darcy and Pipe Turbulence favoring $K_n=12$ and $K_n=8$, respectively. 
These results highlight the importance of choosing an appropriate neighborhood scale.
\begin{figure}[t]
    \centering
    \includegraphics[width=\linewidth]{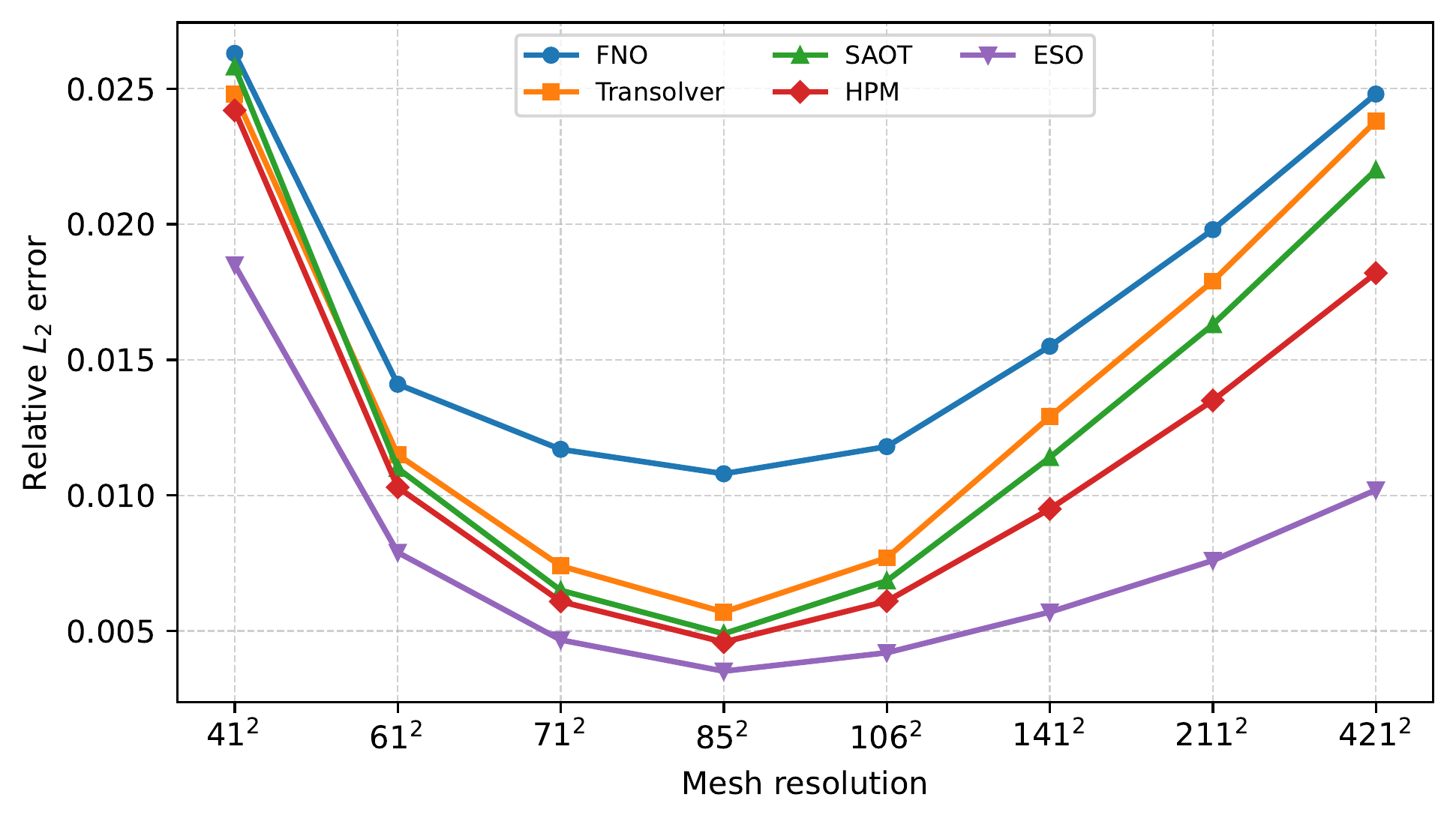}
    \caption{Resolution generalization on Darcy Flow. Relative $L_2$ errors are reported across different mesh resolutions.}
    \label{fig:resolution_generalization}
\end{figure}

\paragraph{Resolution Generalization.}
Figure~\ref{fig:resolution_generalization} compares the resolution generalization ability of different methods on Darcy Flow. 
All methods show a U-shaped trend, where the error first decreases as the mesh becomes finer and then increases at higher resolutions. 
Compared with existing baselines, ESO achieves consistently lower errors across all tested resolutions. 
More importantly, when the resolution increases beyond $85^2$, ESO exhibits a much slower error growth than other methods. 
This indicates that incorporating local pairwise variation helps ESO better preserve fine-scale physics-sensitive structures, thus improving its generalization ability.

\section{Conclusion}

In this paper, we introduced ESO, an edge-conditioned spectral operator for learning PDE solution operators with physics-sensitive local structures. ESO employs PVMM to encode pairwise variations, enabling spectral mode selection to adapt to coefficient jumps, vortical structures, and other local changes while preserving efficient global spectral mixing. We further proposed PAR to emphasize task-dependent sensitive regions during training. Experiments on nine structured and unstructured PDE benchmarks show that ESO consistently outperforms existing neural operators. Qualitative results, ablations, and resolution-generalization analysis further validate the effectiveness of PVMM, PAR, and neighborhood design, suggesting that combining global spectral representations with local edge-aware conditioning is a promising direction for robust neural operators.

\clearpage
\bibliography{aaai2027}

\clearpage
\appendix
\section*{Appendix}
\section{Proofs}

We follow the notation of the main paper. The latent feature at node $x_i$ is written as $h_i$ or $v_t(x_i)$ in the main text. In the implementation details below, $X_i$ denotes the per-head projected latent feature used inside the spectral mixer.

\begin{Theorem}[No-worse approximation]
Let $\mathcal{F}_{\mathrm{base}}$ denote the base point-conditioned spectral operators considered in the main text, and let $\mathcal{F}_{\mathrm{edge}}$ denote the ESO operators with PVMM. For any data distribution $\mathcal{D}$ and any nonnegative loss function $\ell$,
\begin{equation}
    \inf_{f\in\mathcal{F}_{\mathrm{edge}}}
    \mathbb{E}_{(a,u)\sim\mathcal{D}}
    [\ell(f(a),u)]
    \leq
    \inf_{f\in\mathcal{F}_{\mathrm{base}}}
    \mathbb{E}_{(a,u)\sim\mathcal{D}}
    [\ell(f(a),u)] .
\end{equation}
\end{Theorem}

\paragraph{Proof.}
ESO differs from the base spectral operator only in the spectral gate logits. The main paper writes the edge-conditioned gate as a point logit plus an edge-conditioned correction. In the implementation, this correction is parameterized as
\begin{equation}
    g^E_{b,h,i,k}
    =
    g^{\mathrm{base}}_{b,h,i,k}
    +
    \alpha_s \tanh(\alpha_E) b^E_{b,h,i,k},
\end{equation}
where $b^E$ is produced by PVMM, $\alpha_E$ is a learnable scalar, and $\alpha_s$ is a fixed scale. This is the residual-logit implementation of Eq.~(edge gate) in the main paper. Setting $\alpha_E=0$ in code, or equivalently setting the edge coefficient $\lambda_E=0$ in the main text, makes the edge correction exactly zero. The ESO block then becomes the original point-conditioned spectral block under the same basis and spectral transform. Therefore,
\begin{equation}
    \mathcal{F}_{\mathrm{base}}\subseteq \mathcal{F}_{\mathrm{edge}} .
\end{equation}
Taking the infimum of the same nonnegative objective over a larger function class cannot increase the optimal risk, which proves the claim.

\begin{Theorem}[Indistinguishability separation]
Suppose there exist two admissible inputs $a$ and $\widetilde{a}$ and a node $i$ such that
\begin{equation}
    h_i(a)=h_i(\widetilde{a}),
    \qquad
    e_i(a)\neq e_i(\widetilde{a}),
\end{equation}
where $h_i$ is the center representation and $e_i$ is the local edge descriptor. Assume that the target operator satisfies
\begin{equation}
    \left\|
    \mathcal{G}(a)(x_i)-\mathcal{G}(\widetilde{a})(x_i)
    \right\|
    \geq \Delta,\qquad \Delta>0 .
\end{equation}
Then any model whose node-wise prediction at $x_i$ is determined only by $h_i$ must incur pointwise error at least $\Delta/2$ on one of the two inputs. ESO is not subject to this indistinguishability lower bound because its modal weights depend on both $h_i$ and $e_i$.
\end{Theorem}

\paragraph{Proof.}
For a point-conditioned model, the two inputs give the same center representation at node $i$. Hence the model must output the same prediction at this node; denote it by $c$. By the triangle inequality,
\begin{equation}
\begin{split}
    \Delta
    &\leq
    \left\|
    \mathcal{G}(a)(x_i)-\mathcal{G}(\widetilde{a})(x_i)
    \right\| \\
    &\leq
    \left\|
    \mathcal{G}(a)(x_i)-c
    \right\|
    +
    \left\|
    c-\mathcal{G}(\widetilde{a})(x_i)
    \right\|.
\end{split}
\end{equation}
Thus at least one of the two errors is no smaller than $\Delta/2$.

ESO observes the descriptors $(h_i(a),e_i(a))$ and $(h_i(\widetilde{a}),e_i(\widetilde{a}))$, which are distinct because the edge descriptors differ. On these two distinct descriptors, there exists a continuous local response map that assigns the two target values separately. For example, with $c_1=(h_i(a),e_i(a))$, $c_2=(h_i(\widetilde{a}),e_i(\widetilde{a}))$, $y_1=\mathcal{G}(a)(x_i)$, and $y_2=\mathcal{G}(\widetilde{a})(x_i)$, define
\begin{equation}
    \rho(c)=
    \frac{\|c-c_1\|^2}
    {\|c-c_1\|^2+\|c-c_2\|^2},
    r(c)=(1-\rho(c))y_1+\rho(c)y_2 .
\end{equation}
Then $r(c_1)=y_1$ and $r(c_2)=y_2$. Since the edge-bias network in PVMM is implemented as a pointwise network on the local descriptor, using either $1\times1$ convolutions on structured meshes or MLP layers on unstructured meshes, it can approximate such finite descriptor-to-response separation with sufficient hidden width. Therefore, ESO can distinguish the two inputs through $e_i$ and is not forced to use the same node-wise response.

\begin{table*}[t]
\centering
\small
\begin{tabularx}{\textwidth}{C{0.19\textwidth}C{0.29\textwidth}YY}
\toprule
\textbf{Benchmark} & \textbf{PAR field $q$ in code} & \textbf{Neighborhood} & \textbf{Emphasized region} \\
\midrule
Darcy Flow & input permeability coefficient $a$ & structured forward grid jumps & coefficient interfaces \\
Navier--Stokes & next-step vorticity $\omega_{t+1}$ during rollout & structured forward grid jumps & vortices and shear/high-gradient regions \\
Airfoil & airfoil mesh coordinates $(X,Y)$ & structured forward grid jumps & geometry-sensitive near-body regions \\
Plasticity & rigid die-shape input broadcast over the grid & structured forward grid jumps & contact- and geometry-induced deformation regions \\
Irregular Darcy & input coefficient field $c$ & spectral-coordinate kNN graph & coefficient interfaces on triangular mesh \\
Pipe Turbulence & input flow state & spectral-coordinate kNN graph & local velocity-state transitions \\
Heat Transfer & temperature state & spectral-coordinate kNN graph & high-temperature-gradient regions \\
Composite & input thermal-loading field $T$ & spectral-coordinate kNN graph & thermal-loading transitions \\
Blood Flow & velocity trajectory $(v_x,v_y,v_z)$ & spectral-coordinate kNN graph & localized velocity changes over time \\

\bottomrule
\end{tabularx}
\caption{Task-specific PAR fields used by the implementation. Next-step based fields are used only during training to compute loss weights and are not required at inference time.}
\label{tab:appendix-par-q}
\end{table*}

\paragraph{Darcy example.}
In Darcy flow,
\begin{equation}
    -\nabla\cdot(a(x)\nabla u(x))=f(x),
\end{equation}
the local solution depends on fluxes across cell interfaces, which are controlled by permeability contrasts between neighboring cells. Consider two points with the same center permeability $a_i=\widetilde{a}_i$. In one case, the neighbors have similar permeability values; in the other case, one neighbor lies across a sharp material interface. A point-conditioned gate that only observes the center state cannot distinguish these cases. PVMM distinguishes them because the local statistics $D_i$ and especially $Q_i$ become large near the permeability jump. This is a concrete instance of the separation condition above.

\section{PAR Implementation Details}

PAR is implemented as a loss reweighting term. It is separate from PVMM: PVMM uses latent features inside the spectral gate, whereas PAR uses task-available physical fields to construct training weights.

Given a nonnegative sensitivity score $s_{b,i}$, the code normalizes it as
\begin{equation}
    \bar{s}_b=
    \frac{1}{N}\sum_{i=1}^{N}s_{b,i},
\end{equation}
\begin{equation}
    \widetilde{w}_{b,i}
    =
    1+\gamma\frac{s_{b,i}}{\bar{s}_b+\varepsilon},
    \qquad
    w_{b,i}
    =
    \frac{\widetilde{w}_{b,i}}
    {\frac{1}{N}\sum_{j=1}^{N}\widetilde{w}_{b,j}+\varepsilon}.
\end{equation}
Thus the mean weight in each sample is one, which prevents PAR from changing the global loss scale too strongly.

The main paper presents PAR with a weighted squared-error objective for clarity. Since all experiments are evaluated by relative $L_2$ error, the code uses the corresponding weighted relative $L_2$ form:
\begin{equation}
    \mathcal{L}_{\mathrm{PAR}}
    =
    \sum_{b}
    \frac{
    \sqrt{
    \sum_i w_{b,i}
    \|\hat{u}_{b,i}-u_{b,i}\|_2^2+\varepsilon
    }}
    {
    \sqrt{
    \sum_i w_{b,i}
    \|u_{b,i}\|_2^2+\varepsilon
    }} .
\end{equation}
The final training loss is
\begin{equation}
    \mathcal{L}
    =
    \mathcal{L}_{\mathrm{rel}}
    +
    \lambda_{\mathrm{PAR}}\mathcal{L}_{\mathrm{PAR}},
\end{equation}
where $\mathcal{L}_{\mathrm{rel}}$ is the standard relative $L_2$ loss.

\subsection{Structured PAR}

For structured-mesh tasks, PAR uses a forward horizontal/vertical jump indicator. For a task field $q_{b,m,n}\in\mathbb{R}^{C_q}$,
\begin{equation}
    s_{b,m,n}
    =
    \frac{1}{C_q}
    \sum_{c=1}^{C_q}
    \left(
    |q^c_{b,m,n+1}-q^c_{b,m,n}|
    +
    |q^c_{b,m+1,n}-q^c_{b,m,n}|
    \right),
\end{equation}
with invalid boundary terms padded by zero. Darcy Flow uses the same idea through the interface-weight function in the Darcy script; the other structured tasks call the shared structured field-jump utility.

\subsection{Unstructured PAR}

For unstructured-mesh tasks, PAR uses the same spectral-coordinate kNN construction as the graph weight utility. Given a task field $q_{b,i}\in\mathbb{R}^{C_q}$,
\begin{equation}
    s_{b,i}
    =
    \frac{1}{|\mathcal{N}_{K_p}(i)|C_q}
    \sum_{j\in\mathcal{N}_{K_p}(i)}
    \|q_{b,j}-q_{b,i}\|_1 .
\end{equation}
For temporal or multi-channel fields, all non-node dimensions are flattened into $C_q$. The main implementation uses $K_p=8$ unless otherwise stated.

The exact implementation of task field is showen in Table~\ref{tab:appendix-par-q}.

\section{Benchmark and Training Details}

\subsection{Structured Mesh Benchmarks}

\begin{table*}[t]
\centering
\small
\begin{tabularx}{\textwidth}{C{0.16\textwidth}C{0.12\textwidth}C{0.13\textwidth}C{0.18\textwidth}C{0.12\textwidth}Y}
\toprule
\textbf{Benchmark} & \textbf{Train/Test} & \textbf{Resolution} & \textbf{Input $\rightarrow$ Output} & \textbf{Epochs} & \textbf{Main ESO setting} \\
\midrule
Darcy Flow & 1000/200 & $85\times85$ & coefficient $\rightarrow$ pressure & 500 & hidden 128, 8 layers, 8 heads, 128 modes \\
Navier--Stokes & 1000/200 & $64\times64$ & 10 historical vorticity frames $\rightarrow$ 10 future frames & 500 & hidden 256, 8 layers, 8 heads, 128 modes \\
Airfoil & 1000/200 & $221\times51$ & airfoil mesh coordinates $\rightarrow$ Mach number & 500 & hidden 128, 8 layers, 8 heads, 128 modes, late edge/geometry gate on layers 4--7 \\
Plasticity & 900/80 & $101\times31$ & die shape $\rightarrow$ 4-channel deformation over 20 time steps & 500 & hidden 128, 8 layers, 8 heads, 128 modes \\
\bottomrule
\end{tabularx}
\caption{Structured-mesh experimental details following the training scripts.}
\label{tab:appendix-structured-details}
\end{table*}

\begin{table*}[t]
\centering
\small
\begin{tabularx}{\textwidth}{C{0.16\textwidth}C{0.12\textwidth}C{0.13\textwidth}C{0.18\textwidth}C{0.12\textwidth}Y}
\toprule
\textbf{Benchmark} & \textbf{Train/Test} & \textbf{Nodes} & \textbf{Input $\rightarrow$ Output} & \textbf{Epochs} & \textbf{Main ESO setting} \\
\midrule
Irregular Darcy & 1000/200 & 2290 & coefficient $\rightarrow$ pressure & 500 & hidden 128, 4 layers, 8 heads, 32 modes, $K_n=8$ \\
Pipe Turbulence & 300/100 & 2673 & previous flow state $\rightarrow$ next flow state & 1000 & hidden 128, 8 layers, 8 heads, 64 modes, $K_n=8$ \\
Heat Transfer & 100/100 & 7199 & initial/boundary temperature $\rightarrow$ future 3D temperature & 5000 & hidden 32, 4 layers, 1 head, 128 modes, two-domain spectral transition, $K_n=8$ \\
Composite & 400/100 & 8232 & thermal loading $\rightarrow$ deformation & 5000 & hidden 128, 4 layers, 8 heads, 64 modes, $K_n=8$ \\
Blood Flow & 400/100 & 1656 & inlet/outlet boundary conditions $\rightarrow$ blood velocity trajectory & 500 & hidden 64, 4 layers, 8 heads, 64 spatial modes, 16 temporal modes, $K_n=8$ \\
\bottomrule
\end{tabularx}
\caption{Unstructured-mesh experimental details following the training scripts.}
\label{tab:appendix-unstructured-details}
\end{table*}

Darcy Flow uses the standard piecewise-constant Darcy training and testing files, downsampled by a factor of 5. Navier--Stokes uses the first 10 vorticity snapshots as input and predicts the next 10 snapshots autoregressively with step size 1. Airfoil uses the NACA body-fitted coordinate arrays as geometry input and the Mach-number channel as output. Plasticity broadcasts the one-dimensional die-shape input along the second grid axis and predicts deformation fields at continuous queried time values.

\subsection{Unstructured Mesh Benchmarks}

For Irregular Darcy and Pipe Turbulence, the Laplace--Beltrami eigenvectors are computed from the mesh nodes and triangle elements before training. Heat Transfer uses separate precomputed eigenvector matrices for the input and output domains and performs a cross-domain spectral transition. Composite computes mesh spectral features from the provided composite mesh. Blood Flow uses precomputed Laplace--Beltrami eigenvectors for the vessel mesh and additionally performs temporal spectral mixing over 121 time steps.

\subsection{Optimization}

All ESO experiments use AdamW with weight decay $10^{-5}$ and a OneCycle learning-rate schedule. The relative $L_2$ error is used for reporting:
\begin{equation}
    \mathrm{Rel}\text{-}L_2
    =
    \frac{\|\hat{\mathbf{u}}-\mathbf{u}\|_2}
    {\|\mathbf{u}\|_2}.
\end{equation}
Gradient clipping with threshold $0.1$ is enabled in the main scripts where specified. The PAR weight is $\lambda_{\mathrm{PAR}}=0.02$ and $\gamma=2.0$ for most tasks. The airfoil geometry-gated run uses a smaller PAR weight, $\lambda_{\mathrm{PAR}}=5\times10^{-4}$ and $\gamma=0.5$, because the coordinate-based sensitivity is smoother than coefficient jumps in Darcy Flow.

\section{Why PVMM and PAR Are Not Redundant}

PVMM and PAR use local differences for different purposes. PVMM is an architectural module: it computes $D_i$ and $Q_i$ from hidden latent features and uses them to change spectral mode selection. PAR is a training objective: it computes a nonlearned sensitivity weight from a task-available physical field $q$ and increases the loss contribution at sensitive nodes. In particular, structured PVMM uses an 8-neighbor latent stencil with dilation, while structured PAR uses forward physical jumps along the logical grid axes. Thus PAR does not duplicate the PVMM operation; it supplies explicit supervision to regions where accurate prediction is physically important.

\section{More Visualizations}
\begin{figure*}[p]
    \centering
    \includegraphics[width=0.98\linewidth]{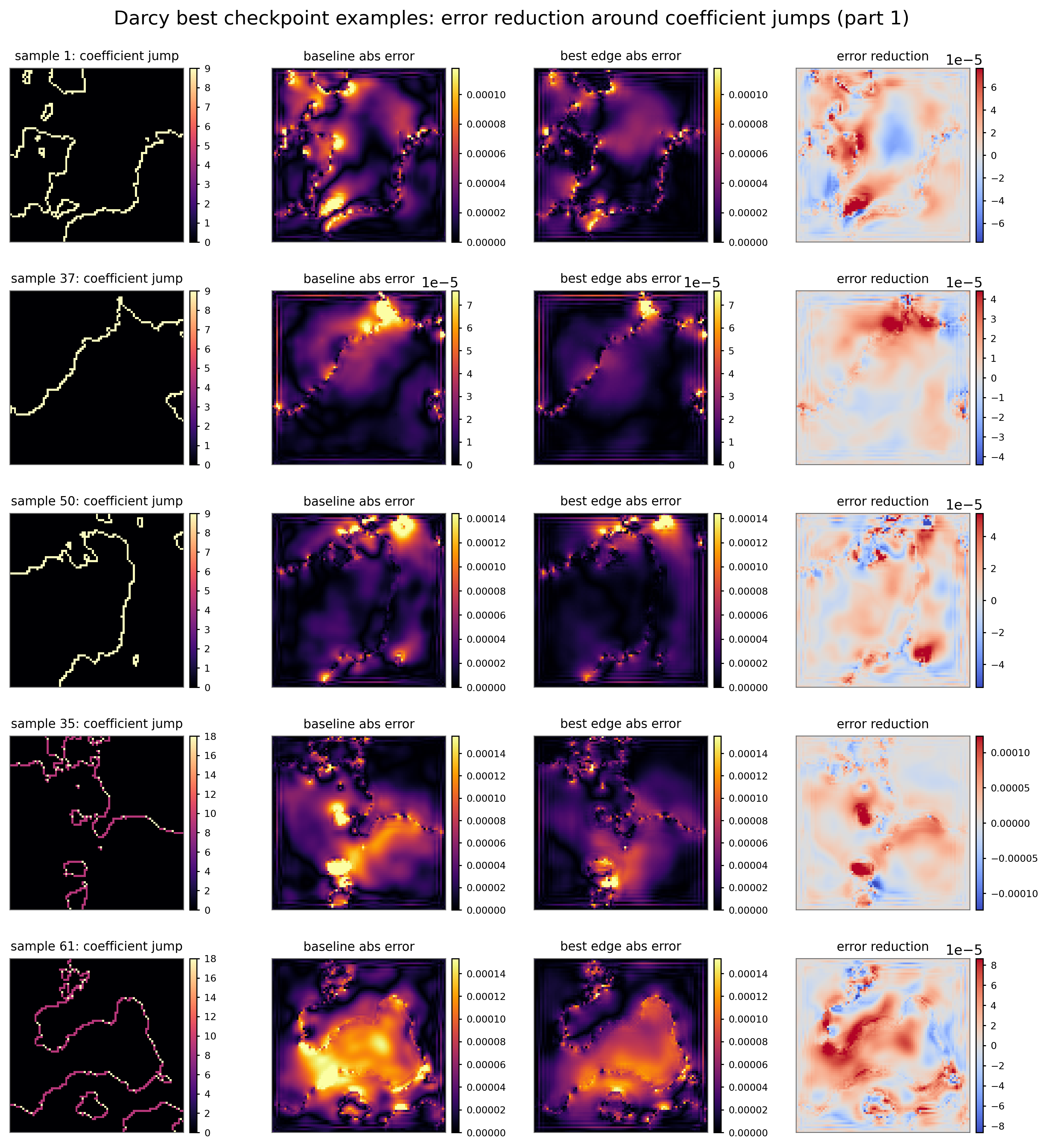}
    \caption{Additional Darcy Flow qualitative examples, part 1.}
    \label{fig:darcy-five-1}
\end{figure*}

\begin{figure*}[p]
    \centering
    \includegraphics[width=0.98\linewidth]{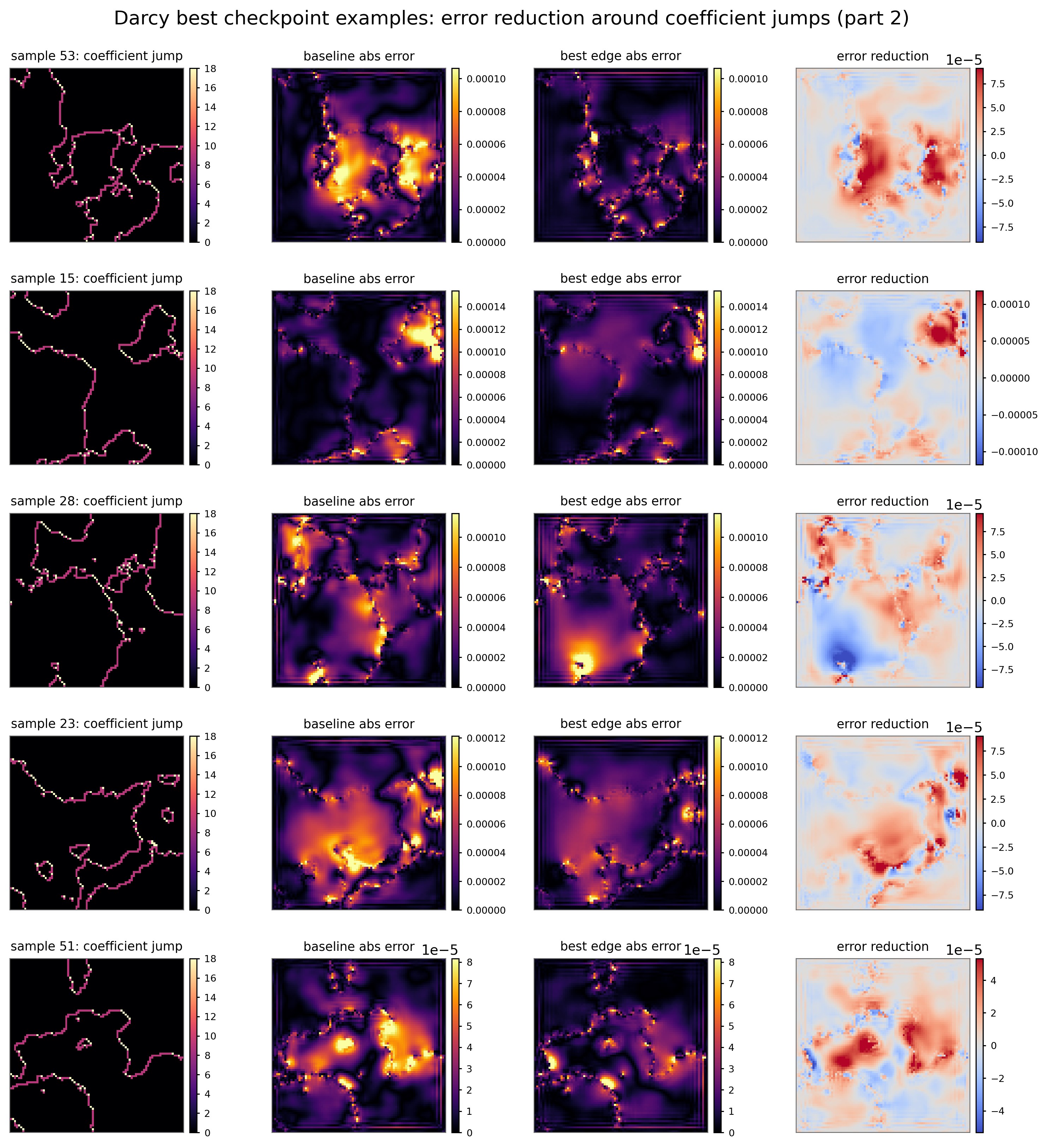}
    \caption{Additional Darcy Flow qualitative examples, part 2.}
    \label{fig:darcy-five-2}
\end{figure*}

\begin{figure*}[p]
    \centering
    \includegraphics[width=0.98\linewidth]{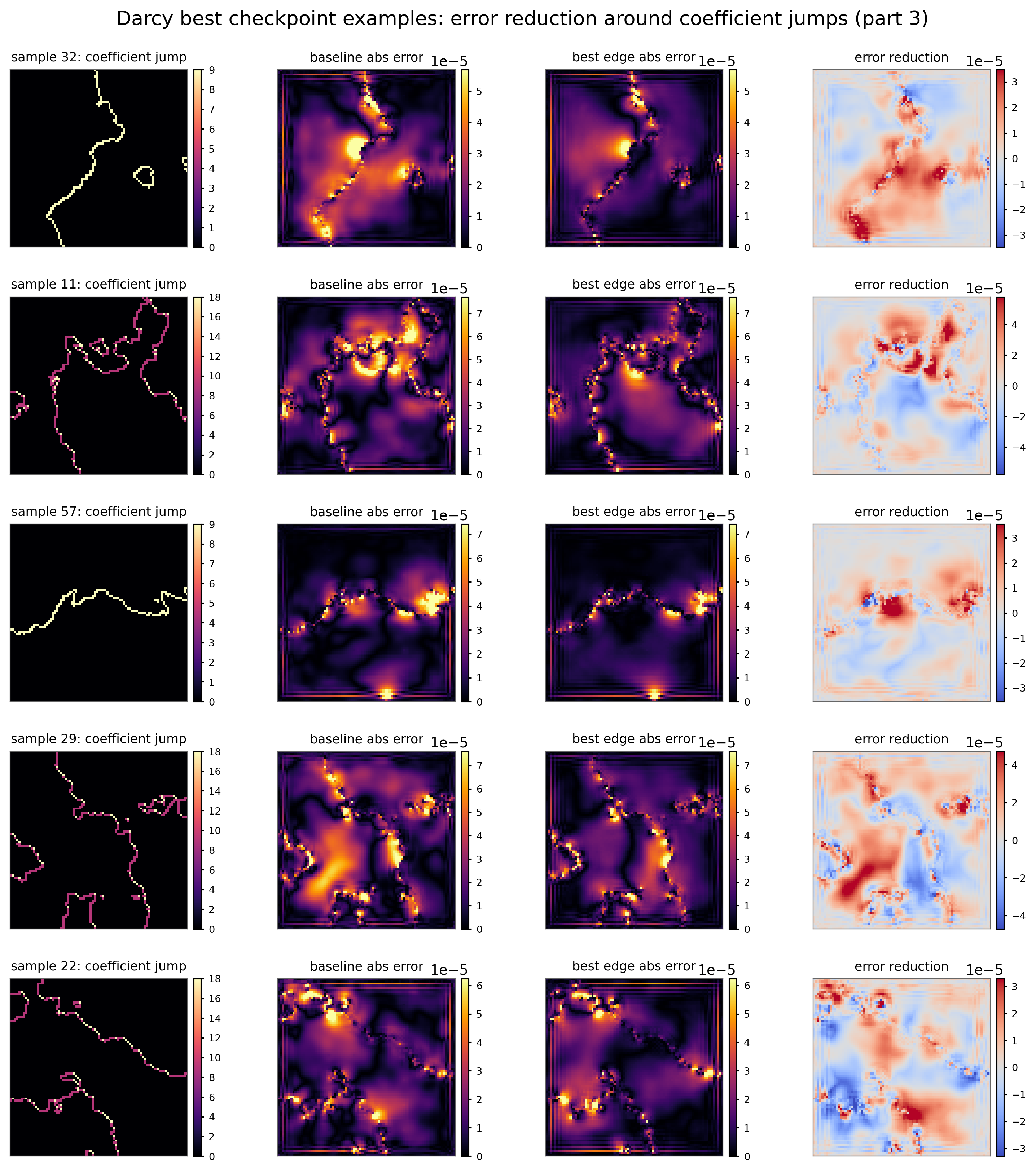}
    \caption{Additional Darcy Flow qualitative examples, part 3.}
    \label{fig:darcy-five-3}
\end{figure*}

\begin{figure*}[p]
    \centering
    \includegraphics[width=0.98\linewidth]{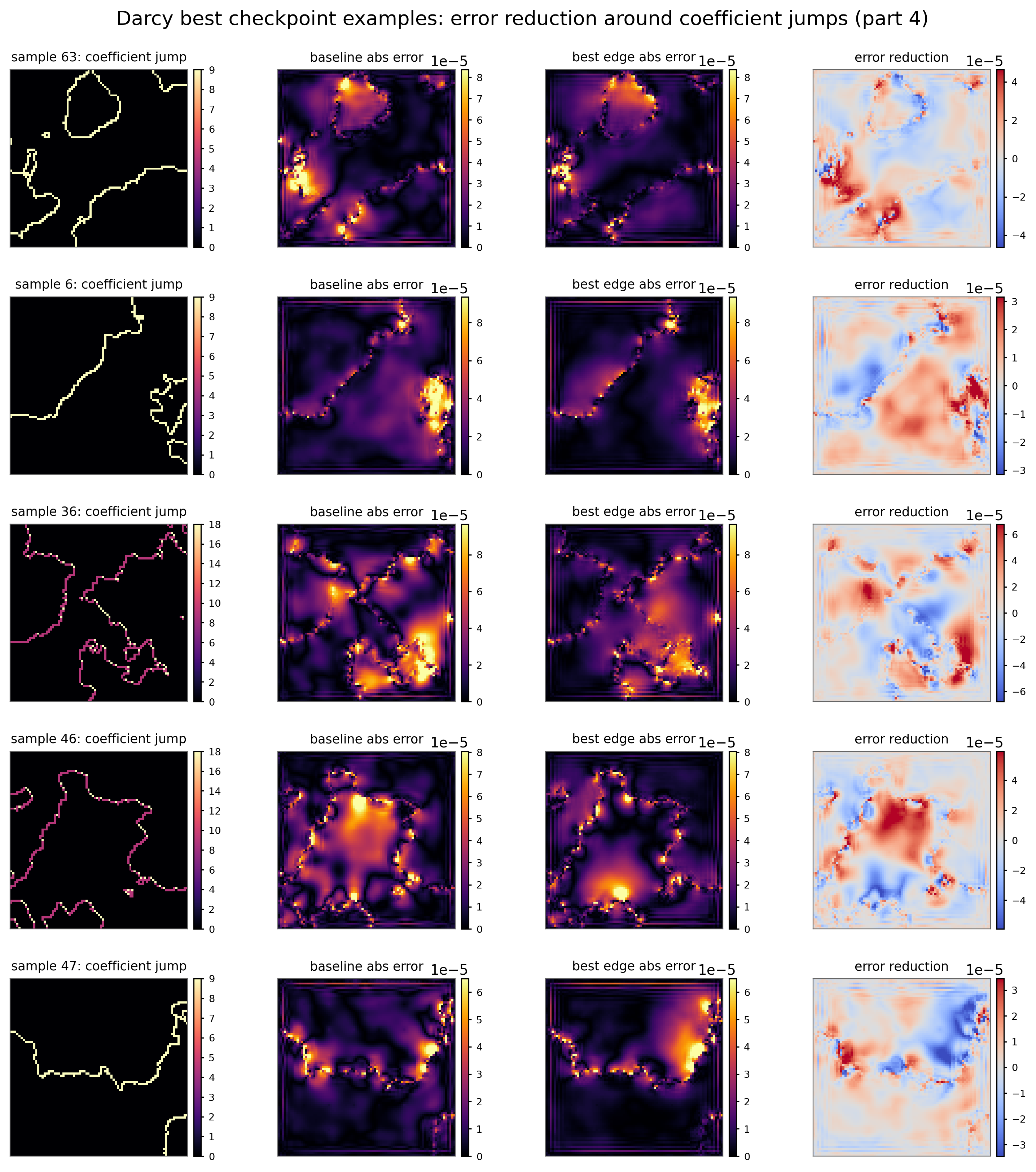}
    \caption{Additional Darcy Flow qualitative examples, part 4.}
    \label{fig:darcy-five-4}
\end{figure*}

\begin{figure*}[p]
    \centering
    \includegraphics[width=0.98\linewidth]{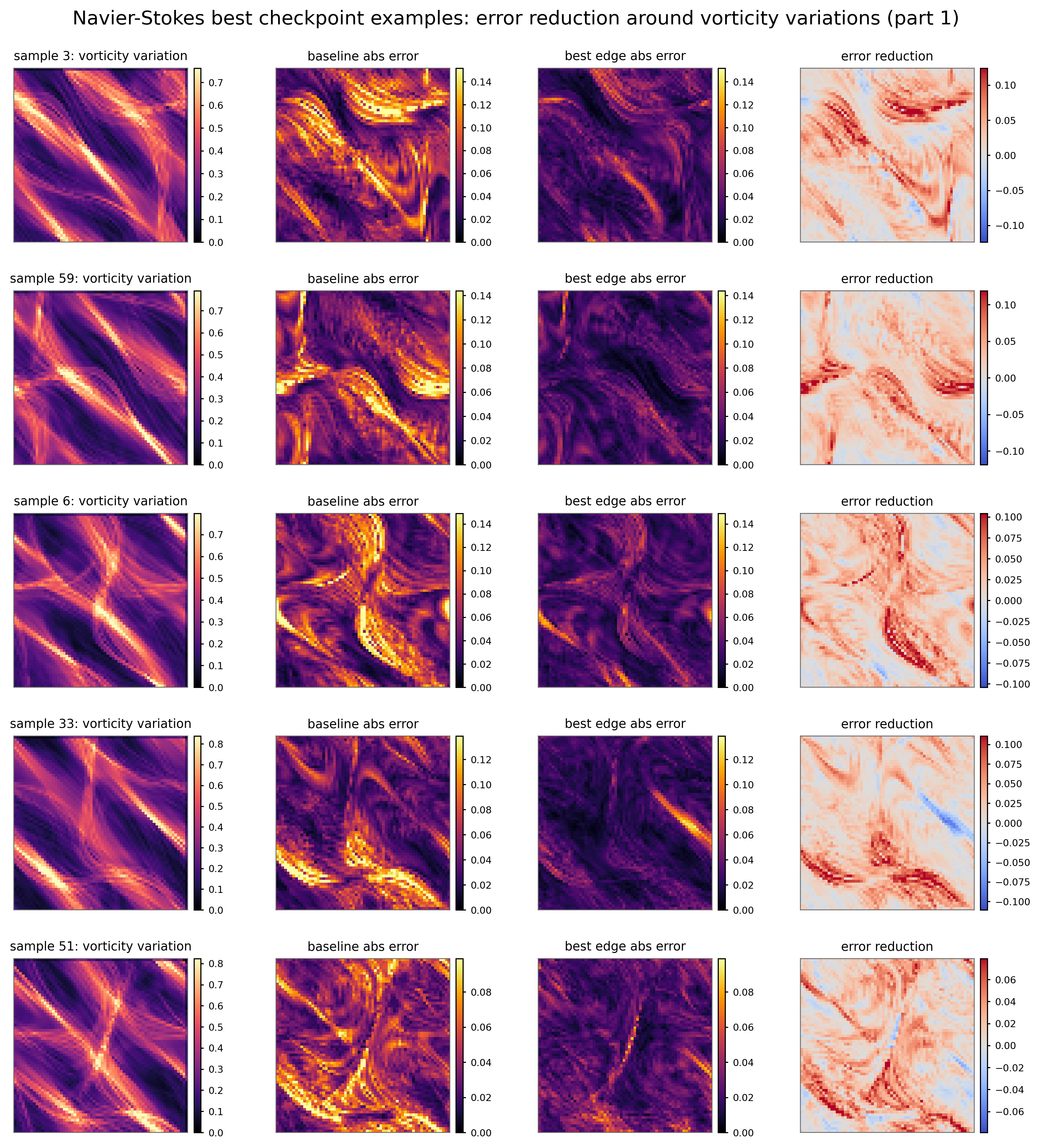}
    \caption{Additional Navier--Stokes qualitative examples, part 1.}
    \label{fig:ns-five-1}
\end{figure*}

\begin{figure*}[p]
    \centering
    \includegraphics[width=0.98\linewidth]{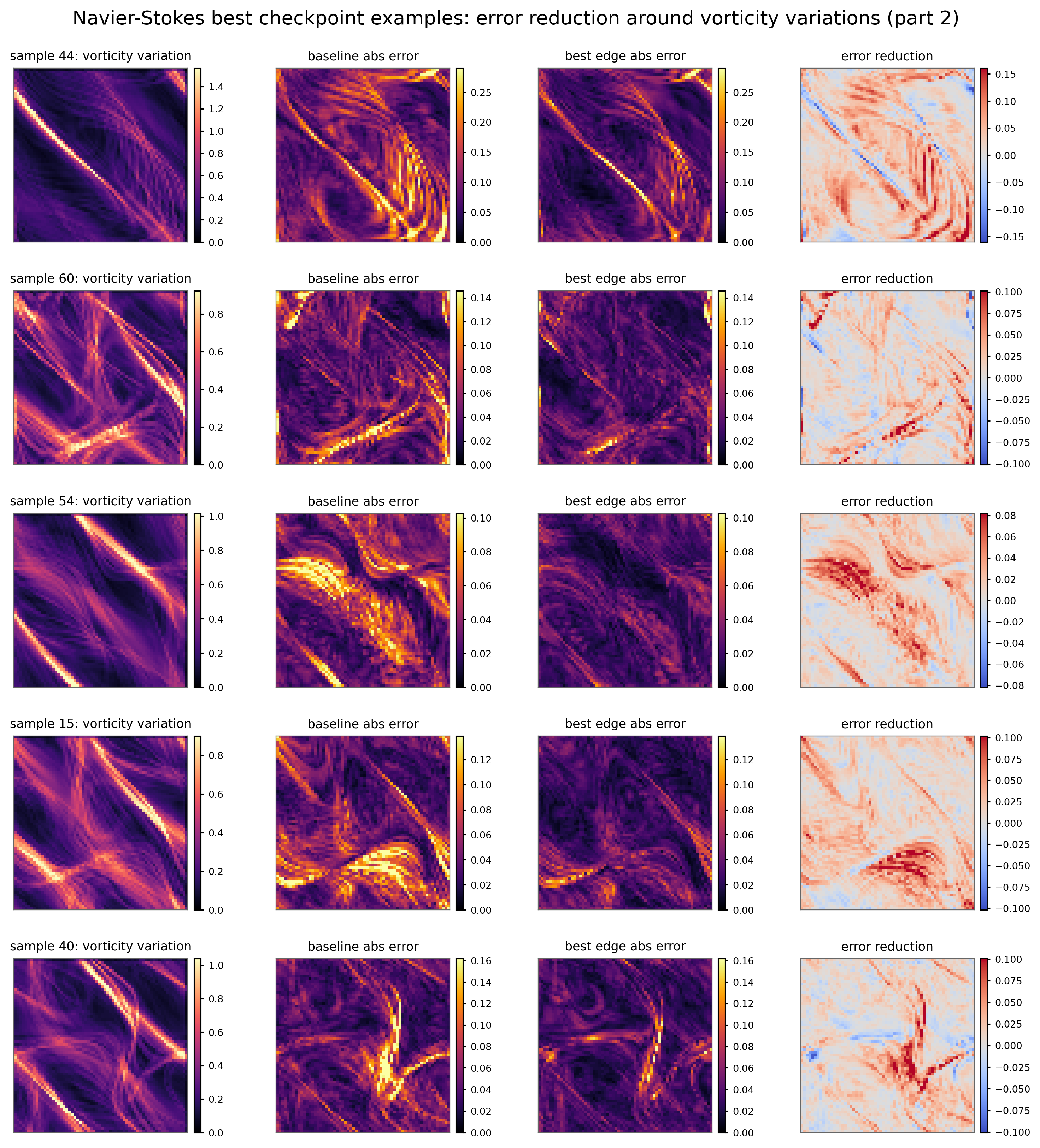}
    \caption{Additional Navier--Stokes qualitative examples, part 2.}
    \label{fig:ns-five-2}
\end{figure*}

\begin{figure*}[p]
    \centering
    \includegraphics[width=0.98\linewidth]{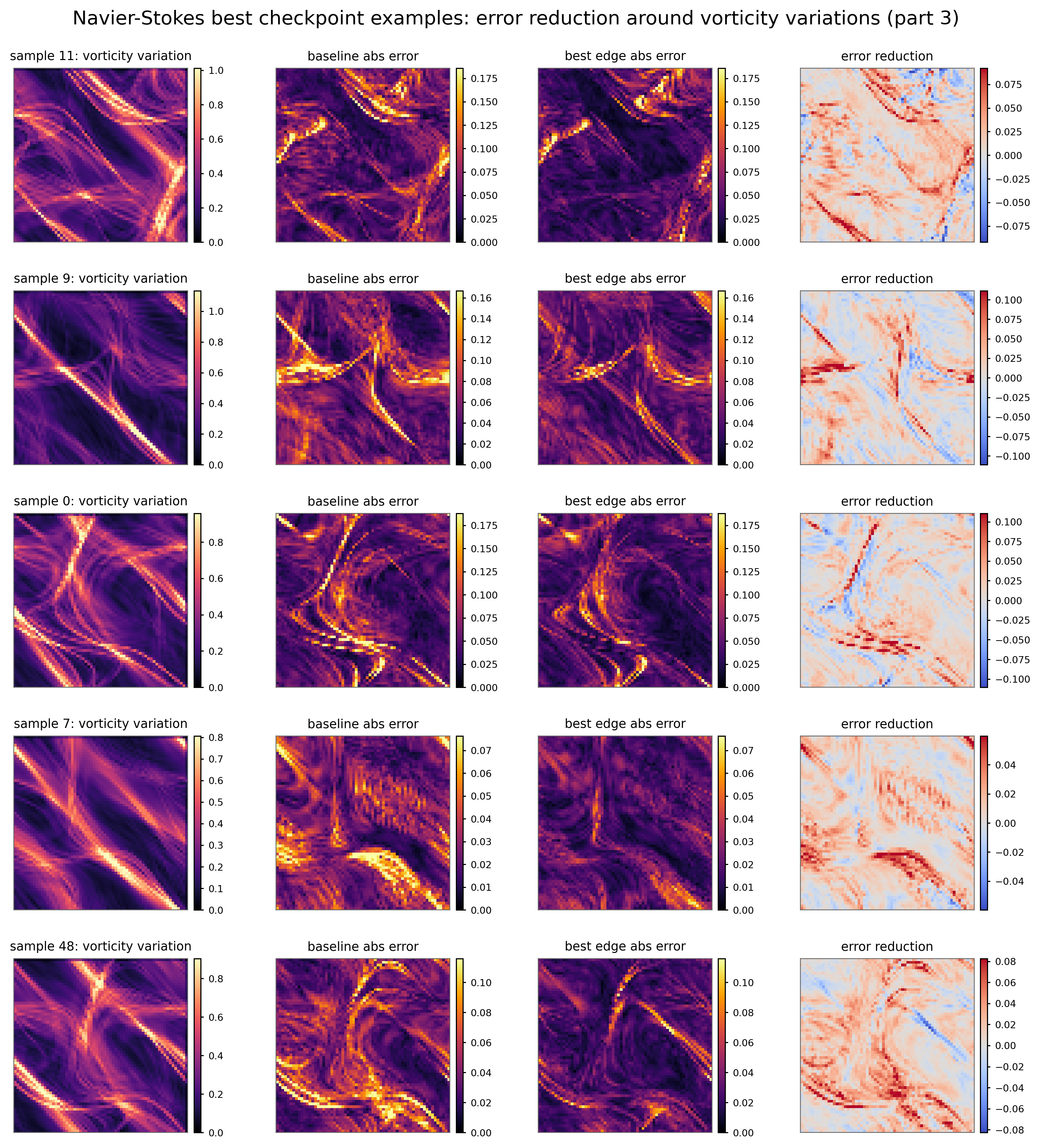}
    \caption{Additional Navier--Stokes qualitative examples, part 3.}
    \label{fig:ns-five-3}
\end{figure*}

\begin{figure*}[p]
    \centering
    \includegraphics[width=0.98\linewidth]{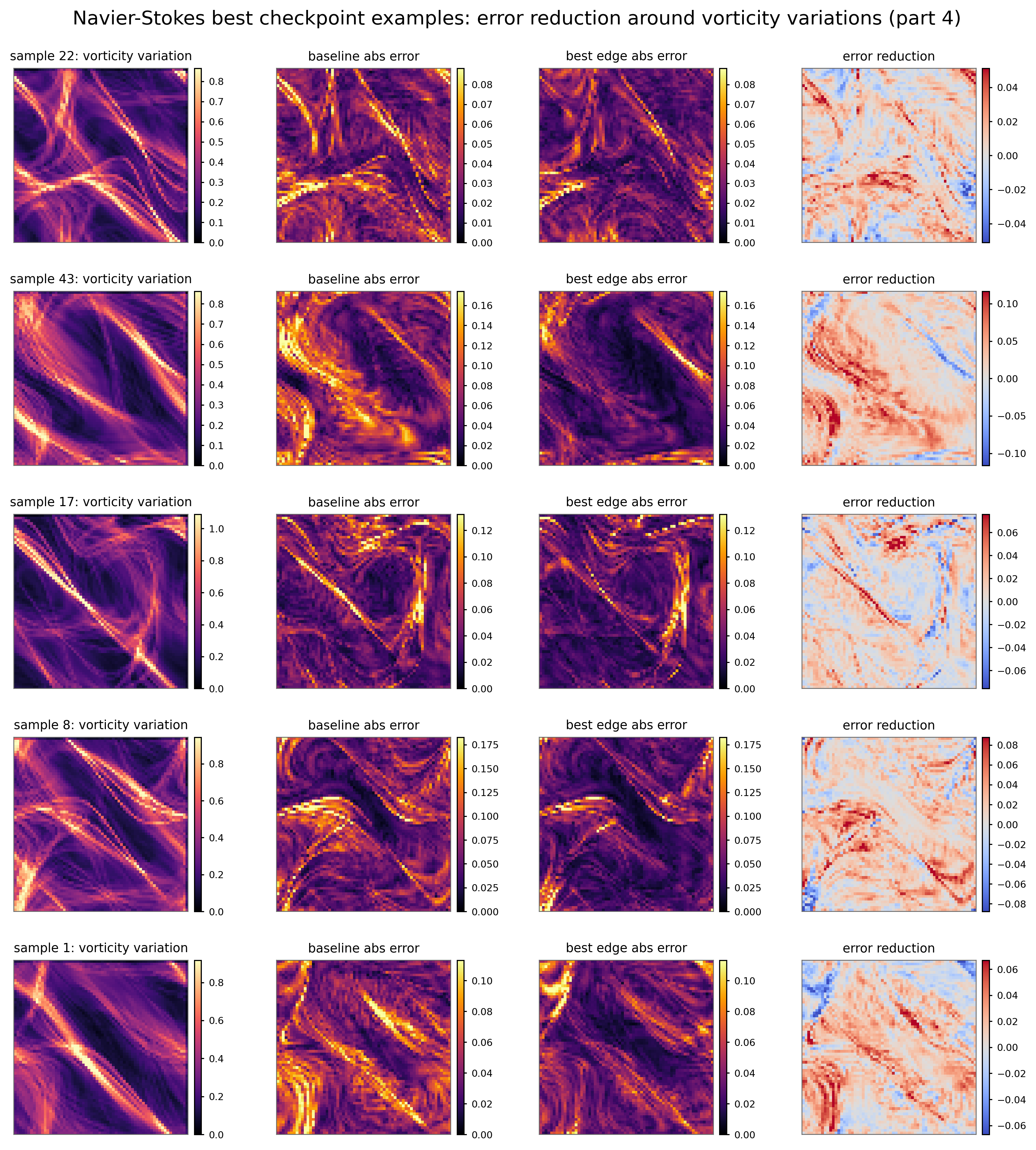}
    \caption{Additional Navier--Stokes qualitative examples, part 4.}
    \label{fig:ns-five-4}
\end{figure*}

Figures~\ref{fig:darcy-five-1}--\ref{fig:darcy-five-4} show additional Darcy Flow examples, with five samples in each figure. Figures~\ref{fig:ns-five-1}--\ref{fig:ns-five-4} show additional Navier--Stokes examples in the same format. In each figure, the four columns are the sensitivity indicator, the baseline absolute error, the ESO absolute error, and the error reduction. Red regions in the error-reduction column indicate locations where ESO has lower error than the baseline.

\end{document}